%% file: main.tex
\documentclass[sn-mathphys,Numbered]{sn-jnl}

\usepackage{fix-cm}
\usepackage[acronym]{glossaries}
\usepackage[utf8]{inputenc}
\makeglossaries

\usepackage{amsmath,amssymb,amsfonts}
\usepackage{amsthm}
\usepackage{mathrsfs}
\usepackage{graphicx}
\usepackage{verbatim}
\usepackage{floatrow}

\usepackage[title]{appendix}
\usepackage{xcolor}
\usepackage{textcomp}
\usepackage{manyfoot}
\usepackage{booktabs}
\usepackage{listings}
\usepackage{multirow}

\usepackage{nicematrix}
\usepackage{color,soul}

\usepackage{ulem} 

\makeatletter
\@ifpackageloaded{nicematrix}{}{%
	\newenvironment{NiceTabular}[1]{\begin{tabular}{#1}}{\end{tabular}}
	\newcommand{\CodeBefore}{}
	\newcommand{\Body}{}
	\newcommand{\Hline}{\hline}
	\newcommand{\Block}[2]{#2}
	\newcommand{\rowcolors}[3]{}
	\newcommand{\rowcolor}[3][]{}
}
\makeatother

\begin{document}

\title[Morphological Characterization of Biological Filaments]{Image-based Morphological Characterization of Filamentous Biological Structures with Non-constant Curvature Shape Feature}

\author*[1,2]{\fnm{Jie} \sur{Fan}}\email{jie.fan.0414@gmail.com}

\author[1,3]{\fnm{Francesco} \sur{Visentin}}

\author[1]{\fnm{Barbara} \sur{Mazzolai}}

\author*[1]{\fnm{Emanuela} \sur{Del Dottore}} \email{deldot@sdu.dk}

\affil[1]{\orgdiv{Bioinspired Soft Robotics Lab}, \orgname{Istituto Italiano di Tecnologia}, \orgaddress{\city{Genova},\country{Italy}}}

\affil[2]{\orgdiv{The BioRobotics Institute}, \orgname{Scuola Superiore Sant'Anna}, \orgaddress{\city{Pisa}, \country{Italy}}}

\affil[3]{\orgdiv{Department of Engineering for Innovation Medicine, Section Engineering}, \orgname{University of Verona}, \orgaddress{\city{Verona}, \country{Italy}}}

\abstract{
Tendrils coil their shape to anchor the plant to supporting structures, allowing vertical growth toward light. Although climbing plants have been studied for a long time, extracting information regarding the relationship between the temporal shape change, the event that triggers it, and the contact location is still challenging.
To help build this relation, we propose an image-based method by which it is possible to analyze shape changes
over time in tendrils when mechano-stimulated in different portions of their body.
We employ a geometric approach using a 3D Piece-Wise Clothoid-based model to reconstruct the configuration taken by a tendril after mechanical rubbing. The reconstruction shows high robustness and reliability with an accuracy of $R^2 > 0.99$. 
This method demonstrates distinct advantages over deep learning-based approaches, including reduced data requirements, lower computational costs, and interpretability.
Our analysis reveals higher responsiveness in the apical segment of tendrils, which might correspond to higher sensitivity and tissue flexibility in that region of the organs. Our study provides a methodology for gaining new insights into plant biomechanics and offers a foundation for designing and developing novel intelligent robotic systems inspired by climbing plants.}


\maketitle

\section{Introduction}\label{section:intro}

In recent years, soft robotics has benefited from studying natural organisms to generate robots with dexterity and compliant structures for better adaptation, prompt response, and safe environment interaction \cite{laschi_2016soft}. In fact, in nature, it is common to find soft, continuum structures to enable, for instance, adaptable manipulation and versatility, e.g., octopus arms \cite{Laschi_2009, mazzolai2019octopus} or elephant trunks \cite{kim_2013soft}.

Looking at the plant kingdom, we can find numerous examples of slender and continuum structures at different levels. In particular, climbing plants show highly adaptive bodies. They build up the main thin stem by growing and elongating cells from their apical extremities, and they flourish in remarkable species-specific attachment strategies to race skywards for light and claim space. These include twining stems, hooks, leaf-bearers, adventitious clinging roots, adhesive pads, and tendril-bearers \cite{Isnard_2009,Burris_2017}. Tendrils, in particular, are specialized, long, slender filamentous organs endowed with extreme sensitivity to mechano-stimulation and the capability to move in a circumnutation-fashion \cite{Chehab_2008} -- i.e., performing large rotations -- while elongating by growth. When they reach a supporting structure, they coil around it, ensuring a tight grasp \cite{Jaffe_1968}. 

Many botanists have investigated the biology and biomechanical properties of tendrils, focusing on their origin and development \cite{Baena_2018_2}, their flexural property \cite{Vidoni_2015}, and their adaptive attachment mechanisms \cite{Burris_2017}. The histological analysis, in situ measurements of biological parameters, and micro-CT scans \cite{Keklikoglou_2021} are the most used techniques to study static morphological properties. On the other hand, studying the dynamics can add to the static analysis information concerning the relationship between time and type of external stimuli and the morphological adaptation of these organs. Such information can help to design and develop novel robotic artifacts with embodied functionalities and intelligence, simplifying their control \cite{giordano2021perspective}. 

However, to enable these types of investigations, it is necessary to develop ad-hoc experimental methodology for observation and analysis. Traditionally, motion tracking and dynamic reconstruction of objects rely on markers placed on the subject's surface to capture its movement within a scene \cite{Moeslund_2006}. Marker-based systems, while accurate, are impractical for dynamic and continuously deforming structures like plants due to their complex 3D growth patterns, surface changes, and obstructions caused by leaves or overlapping plant parts. Recent advancements in deep learning have paved the way for marker-free approaches \cite{Hartley_2004} that leverage vision-based methods, offering a non-intrusive alternative for motion tracking and 3D reconstruction. Techniques like Neural Radiance Fields (NeRF) \cite{Mildenhall_2021_NeRF} enable high-fidelity 3D reconstruction of scenes from multi-view images, learning a volumetric representation that models both geometry and appearance. NeRF has demonstrated exceptional performance for such deformable objects. Furthermore, methods such as Gaussian Splatting \cite{Kerbl_2023_gaussians} have improved computational efficiency and reconstruction quality, providing real-time capabilities for dynamic 3D scenes. These marker-free approaches have successfully captured complex motions without relying on physical markers.

While learning-based frameworks like NeRF \cite{Mildenhall_2021_NeRF} and Gaussian Splatting \cite{Kerbl_2023_gaussians} excel at photorealistic rendering, they rely on densely sampled viewpoints—often dozens to hundreds of images—to adequately constrain the high-dimensional optimization landscape of neural representations or Gaussian primitives. Their inherent appearance-centric nature optimizes volumetric radiance fields or spherical harmonics to maximize photometric consistency across views, yielding underlying geometric representations that are implicitly encoded and often metrically imprecise, particularly for thin structures. Although recent advances such as GS2Mesh \cite{wolf_2024_gsmesh_eccv}, NeuSG \cite{wang2021neus}, and MonoGSDF \cite{Yu2022MonoSDF} attempt to impose additional geometric regularization or surface extraction from Gaussian primitives, they primarily target smooth surface reconstruction rather than sub-millimeter filamentous morphology. Even approaches explicitly designed for edge or curve recovery, such as the Spherical Gaussian Curve Reconstruction (SGCR) framework \cite{yang2025sgcr}, require large structured multi-view datasets and edge supervision to fit parametric curves, thus remaining unsuitable for unconstrained single- or few-view scenarios.

Furthermore, extensive neural training reduces the suitability of these methods for fine-scale geometric reconstruction of evolving tendril-like structures, since they typically prioritize visual fidelity over precise geometric accuracy. Plant tendrils display extreme aspect ratios (length-to-diameter exceeding 100:1), sparse textures, and semi-transparent or specular surface qualities depending on their developmental stage. These traits violate important assumptions of neural rendering approaches: learning-based methods struggle with textureless areas where view-dependent effects do not provide enough training signals, and they may generate geometrically implausible reconstructions despite photometric accuracy. The thin topology causes sampling ambiguity in voxel or point-based neural representations, where limited occupancy along slender structures results in noisy or incomplete geometry. 

Significant challenges remain when reconstructing growing and deformable continuum bodies such as plant stems, vines, and roots. Unlike rigid or piece-wise rigid bodies, plant structures exhibit non-uniform curling behaviors with smooth transitions between straight and curved segments or between multiple curves. Such behaviors are inherently dynamic and cannot be effectively modeled using conventional constant curvature or piece-wise constant curvature approximations widely adopted in continuum robotics \cite{Webster_2010, Trivedi_2008}. The non-constant curvature character complicates the reconstruction process, as the geometric transitions must be captured with high precision in 3D space while accounting for global growth and local deformations. 

To address this gap, we previously proposed a method based on 2D clothoid spirals (Euler curves) \cite{Fan_2020_1} and Piece-Wise Clothoid (PWC) \cite{Fan_2020_2} to accurately approximate curling shapes in planar space. These approaches demonstrated the ability to model smooth curvature transitions, making them particularly effective for capturing the dynamic behavior of curling and deformable structures. Building on these works, this paper aims to extend the analysis domain from 2D to 3D, addressing the complexities of slender, growing shapes with non-constant curvature in three-dimensional space. Specifically, we focus on extracting precise morphological descriptions and performing accurate 3D reconstructions of filamentous continuum bodies, laying the foundation for new tools in bioinspired robotics and plant biomechanics. This work provides a critical step toward enabling marker-free, high-fidelity tracking of dynamic, deformable systems in real-world environments.

Furthermore, our approach handles complex filamentous structures, including those that self-intersect or occlude portions of their own geometry. By leveraging multi-view images, the method allows for robust and precise reconstruction of intricate 3D shapes, capturing their continuous and evolving morphology over time. This advancement is particularly relevant for tendril-like structures, where mechanical forces and environmental factors influence growth. We validated the proposed reconstruction methodology through experiments on tendrils, where we analyzed their morphological evolution under a constant force applied at different locations along their length. The experiments demonstrated the effectiveness of our approach in accurately capturing and modeling the dynamic changes in tendril shapes, enabling insights into their growth behaviors. Hence, our parametric, geometry-driven approach offers controllability, interpretability, and physical meaning for morphological characterization.

The key contributions of this paper are as follows:

\begin{itemize}
\item at the engineering level
    \begin{itemize}
        \item Developed a novel vision-based method to reconstruct 3D spatial shapes of filamentous continuum structures based on the traditional Multi-View Stereo technique, 
        overcoming challenges such as reduced feature visibility, self-occlusions, and complex deformations.
        \item Proposed a 3D Principal Curve Deformation (PWC) spiral model to mathematically describe the intricate morphology of slender, filamentous, continuum shapes.
    \end{itemize}
\item at the biological level
    \begin{itemize}
        \item introduces a practical tool that can be adopted to characterize the morphological evolution of tendril-like elements over time; 
        \item analyzes the growth patterns to reveal the uneven distribution and transduction of the tendrils' sensitivity along its length.
    \end{itemize}
\end{itemize}

The remainder of this paper is organized as follows. Section \ref{section:mm} introduces the setup of biological experiments designed for morphology observation and data collection. This section also details the proposed shape reconstruction algorithms, covering video pre-processing, skeleton ordering, and point correspondence identification. Additionally, we provide the mathematical definition of the PWC spiral used to model and reconstruct the generated 3D point cloud. Section \ref{section:Results} presents the experimental results, including an analysis of the morphological and temporal evolution of tendrils under applied forces. Finally, Section \ref{section:Discussion} concludes the paper with a discussion of the findings and outlines directions for future work.

\section{Materials and Methods}
\label{section:mm}

\subsection{Plant Growth Conditions}
We tested the proposed three-dimensional reconstruction methodology by analyzing the curling behavior in tendrils of a \textit{Passiflora caerulea}. The plant was purchased from a local flower store during winter, and all the experiments were carried out in the same season. The plant was kept in a growing chamber at $23^\circ$C, with $70\%$ relative humidity and a day/night cycle of $12/12 $ hours. Since matured tendrils can show maximum irritability and little change in their mechano-sensitivity, we chose tendrils in their maturity phase, having a length of about $14.8 - 24$~cm.

\subsection{Experimental Setup}\label{chapter:experiment}
\begin{figure}[htb]
\centering
\includegraphics[width=\textwidth, trim={0 13cm 0 0},clip]{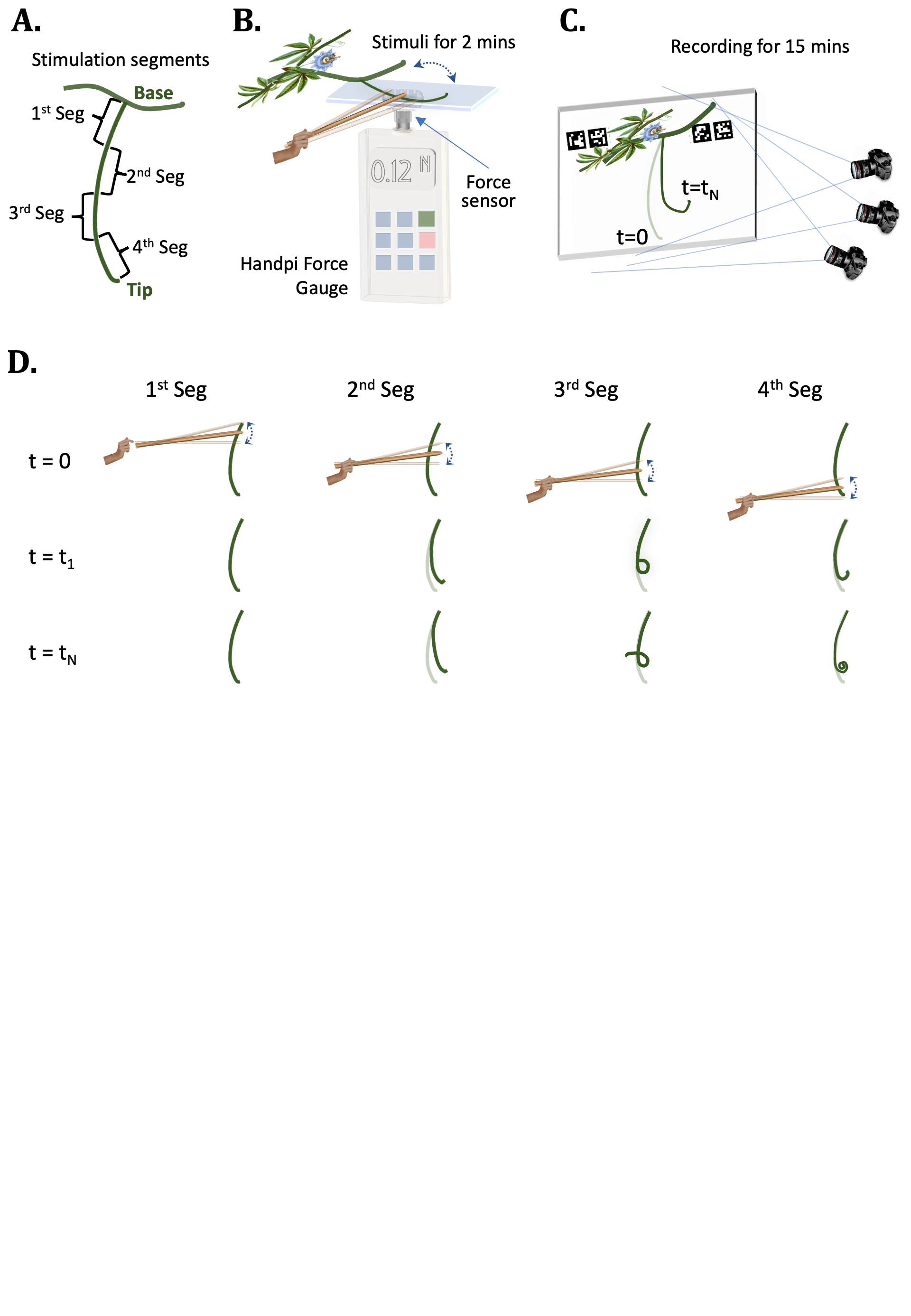}
\caption{Biological experiment setup and data collection procedure. $\mathbf{A}.$ Definition of the portions of the tendril stimulated in the experiments. In this work, we subdivided each tendril into four segments and stimulated only one using a wooden stick. $\mathbf{B}.$ Measurements of the applied force during the experiments. To verify the intensity of the applied stimulus, we measured the interaction with a precision load cell and replicated the same motion over the selected portion for all the tendrils. $\mathbf{C}.$ Experimental setup. After the stimulation, the plant main stem was taped on a panel, leaving the tendril hanging free. On the panel, markers are used to facilitate the data processing and identify common features in the three views used for the 3D reconstruction. $\mathbf{D}.$ Schematic representation of the stimuli-induced morphological changes that occurred over time in the four experimental sets.}
\label{fig:expri_setup}
\end{figure}

In our previous studies \cite{Fan_2020_1, Fan_2020_2}, we focused only on the two-dimensional analysis of the morphology change in the tendril upon contact occurring in a single area. 
Here, we extend to the 3D case and analyze the effects of a constant force applied on different portions along the length of the tendril. It is known that a tendril responds more to a higher frequency of ventral stimuli than to a lower one, although the number of stimuli remains the same \cite{Jaffe_1970}, and that the sensitivity of a tendril is not evenly distributed along its body and decreases as the distance from the tip increases \cite{Jaffe_1966,Jaffe_1970}. Thus, we divided the tendril into four stimulation portions (namely \textbf{S1}, \textbf{S2}, \textbf{S3}, and \textbf{S4}) whose located length is evenly divided along the total length of a tendril. The first portion starts from the stem (base of the tendril), whereas the last portion is close to the tendril apical extremity (Figure~\ref{fig:expri_setup}A). One portion corresponds to one experimental set. We repeated the same experimental procedure in each portion using four different tendrils from the same plant for a total of sixteen experiments divided into four groups (Figure~\ref{fig:expri_setup}A).

We verified the force applied by laying down the tendril to one face of a glass slide and adhering to the other side a digital Force Gauge (Handpi Force Gauge HP-50) equipped with a force sensor (with $0.01$~N resolution, $10$~N capacity) (Figure~\ref{fig:expri_setup}B). 

Then, we continuously stimulated back and forth the selected segment for 2 minutes (frequency: $f = 40~\mathrm{times/min}$, stroking force: $F = 0.12$~N) using a wooden stick. 
After the stimulation, the tendril was left hanging free. The adjacent main stem was secured on a whiteboard (Figure~\ref{fig:expri_setup}C) with a tape to limit possible oscillations and interference introduced by the motion of the stem that can cause unwanted contact with other parts of the plant. In addition, to limit other possible environmental stimuli affecting the results, we performed the experiments in the same growing chamber, keeping constant environmental settings. 

The response of the tendril is quite rapid, already showing initial movements a few seconds after the stimulus. However, the tendril might take longer to complete its motion, depending on the contact location. Thus, we acquired 15 minutes of video ($30$~fps) to cover the whole motion in each experiment. Three cameras (Pentax WG-\uppercase\expandafter{\romannumeral3} with Lens $25-100$~mm, image size is $1920 \times 1080$ pixels) were placed in different, fixed positions having a relative angle of  $60^\circ$ and at a distance from the tendril of $50$~cm. To compute both the camera intrinsic parameters matrix, $\mathbf{K}$, and camera extrinsic parameters matrix, $\left[ \mathbf{R}|\mathbf{t} \right]$, we used multiple ArUco markers close to the tendril under observation to match the three views by the homography transformation (Figure~\ref{fig:expri_setup}C). 

A schematic representation of all the cases and relative outputs are depicted in Figure~\ref{fig:expri_setup}D. From preliminary experiments, we observed that the morphological changes of tendrils in response to touch could be divided into two categories: one producing curling on a plane (stimulus applied in segment \textbf{S1}, \textbf{S2}, and \textbf{S4}), and a second category that generates an off-plane curling (stimulus applied in segment \textbf{S3}), which visually generates a sort of cross-intersection of the tendril (Figure~\ref{fig:expri_setup}D, third column, bottom row). This initial observation guided the development of the algorithm used for the 3D reconstruction.

\subsection{Pre-processing}

\begin{figure}[htb]
\centering
\includegraphics[width=\textwidth]{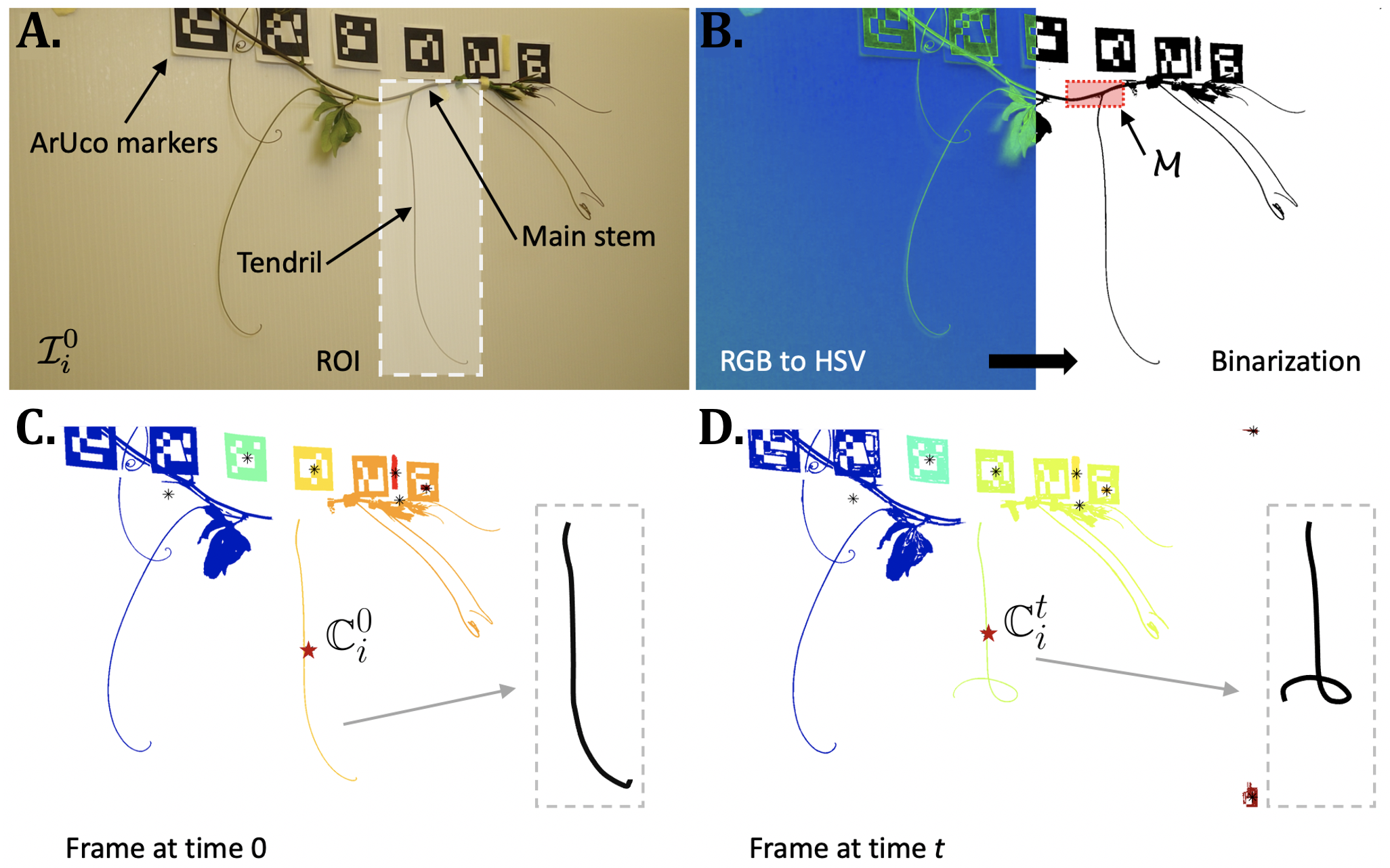}
\caption{Steps of the pre-possessing applied to each captured video. $\mathbf{A}.$ The first frame of each video, $\mathcal{I}^0_i$ ($i=1, 2, 3$), is processed in a semi-automatic fashion. The user is asked to select a ROI that contains the element to be analyzed (i.e., a tendril). The ArUco markers are attached to the background, serving as global correspondence between the images captured by the three cameras and retrieving the camera parameters. $\mathbf{B}.$ The RGB image is then converted to HSV color space (left) and thresholded to extract the foreground (i.e., the selected tendril) from the background (right). Again, we ask the user to select the connection point between the main stem and the tendril to avoid the segmentation algorithm further selecting points not belonging to the tendril. $\mathbf{C}.$ Using the connected components algorithm, each set of pixels is labeled according to its connections with the neighboring ones. Then, the centroid, $\mathbb{C}^0_i$ ($i$ is the $i$-th camera), for the pixels contained in the ROI is computed and stored to automate the extraction of the pixels belonging to the tendril in successive frames. $\mathbf{D}.$ The results of the automatic extraction of the tendril using the updated centroid, $\mathbb{C}^{t-1}_i$ (where $t-1$ is the time frame and $i$ is the $i$-th camera), and considering only the portion of the image contained in the ROI defined in $\mathcal{I}^0_i$.}
\label{fig:segmentation}
\end{figure}

Once cameras are calibrated and three different perspectives (namely $\mathcal{V}_1$, $\mathcal{V}_2$, and $\mathcal{V}_3$) are acquired, we need to match the correspondence between images, extract the skeleton points, and reconstruct a three-dimensional model of the tendril. 
As a pre-processing step, we have to extract the Region Of Interest (ROI) that contains the tendril from each frame of the video, $\mathcal{I}^t_i$ ($i=1, 2, 3$; $t=0,\ldots,m$), and isolate it from the background. The problem is not trivial since the scene contains stems, leaves, and other tendrils. For this reason, a semi-automatic procedure has been developed: the user selects the ROI in the initial frame $\mathcal{I}^0_i$ of each $\mathcal{V}_{i = 1,2,3}$. Then, the user is asked to draw an additional rectangular region between the tendril and the main stem at the intersection point to prevent the algorithm from selecting undesired pixels from the stem. The segmentation process uses this area as a keep-out mask ($\mathcal{M}$) for all the successive frames ($\mathcal{I}^t_i$ ($i=1, 2, 3$; $t=1,\ldots,m$)). The image is then converted from RGB to HSV color space to highlight the difference between background and foreground and binarized by thresholding. The segmentation is performed by computing the connected components, which returns several sets of pixels and their relative labels. Only the components in the ROI are maintained, and their centroids, $\mathbb{C}^0_i$, are computed. The centroid in all the following frames ($t=1,\ldots,m$) is computed by minimizing the distance from the previous, $\mathbb{C}^{t-1}_i$, using K-nearest neighbor search. Figure~\ref{fig:segmentation} depicts the steps involved in the pre-processing of the data, wherein the ROI has only one connected component, and consequently, one centroid is obtained.

\subsection{Ordering the Points} 
The pre-processing phase returns a set of unordered pixels representing the skeleton of the tendril. To reconstruct the shape, they must be ordered. We noticed that the change in the morphology of the tendril either happens in-plane or off-plane. In the first case, we already proposed a solution \cite{Fan_2020_1} in which, starting from a point, the algorithm navigates in two opposite directions and then merges the two ordered subsets. For the off-plane case, the problem is that the tendril might self-occlude, making the sorting more challenging. In the following, we present the algorithm to address this issue.

\begin{figure}[htb!]
\centering
\includegraphics[width=\textwidth]{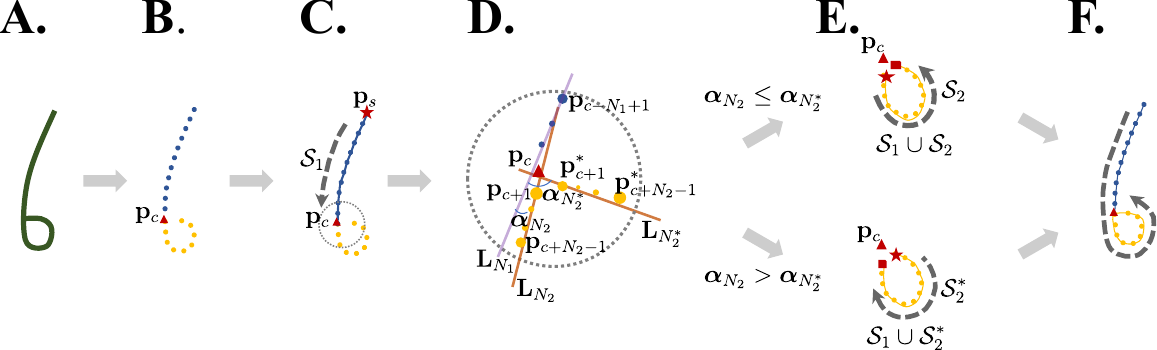}
\caption{Illustration of the sorting algorithm steps for off-plane morphing where the distance between the self-intersection point and tendril tip is negligible. $\mathbf{A}.$ Example of tendril configuration. $\mathbf{B}.$ The skeleton points are split into two sets divided by the self-intersection point ($\mathbf{p}_{c}$). $\mathbf{C}.$ Ordering the first set, $\mathcal{S}_1$, up to the self-intersection point, which is removed. The remaining points are collected in a new set with two ordering directions: clockwise ($\mathcal{S}^*_2$) and counterclockwise ($\mathcal{S}_2$).  $\mathbf{D}.$ For each set, we fit a line using $N_i$ points ($i =1,2$). $\mathbf{E}.$ The correct ordering direction is selected by considering the angles between the lines pairs $\langle\mathbf{L}_{N_1}, \mathbf{L}_{N_2} \rangle$ and $\langle\mathbf{L}_{N_1}, \mathbf{L}_{N^*_2} \rangle$. $\mathbf{F}.$ Final, ordered set of points. }
\label{fig:cross-intersectionA}
\end{figure}

For the off-plane case, the algorithm first checks if there is a self-intersection point by computing the perimeter of the center pixel within a $3 \times 3$ neighborhood filter. If the filter returns an intersection point, the algorithm checks the distance between the tip of the tendril -- i.e., the most apical region -- and the point found.

If the distance is negligible (Figure~\ref{fig:cross-intersectionA}), we can find and remove the cross-point by applying edge linking and line segment fitting from \cite{KovesiMATLABCode} (Figure~\ref{fig:cross-intersectionA}B). Then, we can apply our sorting algorithm starting from the initial point $\mathbf{p}_s$ (marked by a red star in Figure~\ref{fig:cross-intersectionA}) to the cross-point $\mathbf{p}_c$ (Figure~\ref{fig:cross-intersectionA}C). The first set of points is denoted by $\mathcal{S}_1$. For the remaining points, from $\mathbf{p}_c$ to the tip (denoted as $\mathcal{S}_2$), the algorithm can navigate the points either clockwise or counterclockwise (Figure~\ref{fig:cross-intersectionA}E). Both sorting directions can be selected since one is the reverse of the other; however, only one is correct since it generates a smoother curve. We call the two sets $\mathcal{S}_2$ and $\mathcal{S}^*_2$ (where $^*$ indicates the reverse ordering).
After ordering the two sets independently, to identify the correct direction, we select the last $N_1$ points from the set $\mathcal{S}_1$, i.e. $\left\lbrace \mathbf{p}_{c-N_1+1}, \mathbf{p}_{c-N_1+2}, \cdots, \mathbf{p}_{c} \right\rbrace$, and the first $N_2$ and $N^*_2$ points from $\mathcal{S}_2$ and $\mathcal{S}^*_2$ respectively, i.e., $\left\lbrace \mathbf{p}_{c}, \mathbf{p}_{c+1}, \cdots, \mathbf{p}_{c+N_2-1} \right\rbrace$, and $\left\lbrace \mathbf{p}_{c}, \mathbf{p}^*_{c+1}, \cdots, \mathbf{p}^*_{c+N_2-1} \right\rbrace$. The points are used to fit three straight lines, denoted as $\mathbf{L}_{N_1}$, $\mathbf{L}_{N_2}$ and $\mathbf{L}_{N^*_2}$, belonging to each of the three sets of points. By computing the angles between the line pairs $\langle\mathbf{L}_{N_1}, \mathbf{L}_{N_2} \rangle$ and $\langle\mathbf{L}_{N_1}, \mathbf{L}_{N^*_2} \rangle$:

\begin{equation}
\displaystyle
	\begin{split}
    	\boldsymbol{\alpha}_{j} &=\rm{tan}^{-1}\left(\rm{det}\left(\left[\overrightarrow{\mathbf{L}_{N_1}};\overrightarrow{\mathbf{L}_{j}}\right]\right), \rm{dot}\left(\overrightarrow{\mathbf{L}_{N_1}},\overrightarrow{\mathbf{L}_{j}}\right)\right), ~~ j=N_2,N^*_2,
	\end{split}
\end{equation}

we can identify the correct sorting direction $\mathcal{S}$ of the whole tendril by comparing the two angles:
\begin{equation}
\displaystyle
\begin{split}
	\mathcal{S} = \left\{
    		\begin{split}
         	& \mathcal{S}_1 \cup \mathcal{S}_2 , \quad if ~ \boldsymbol{\alpha}_{N_{2}} \leq \boldsymbol{\alpha}_{N^*_2} \\
         	& \mathcal{S}_1 \cup \mathcal{S}^*_2, \quad if ~ \boldsymbol{\alpha}_{N_2} > \boldsymbol{\alpha}_{N^*_2}.
    	\end{split}
    	\right.
	\end{split}
\end{equation}

\begin{figure}[htb!]
\centering
\includegraphics[width=\textwidth]{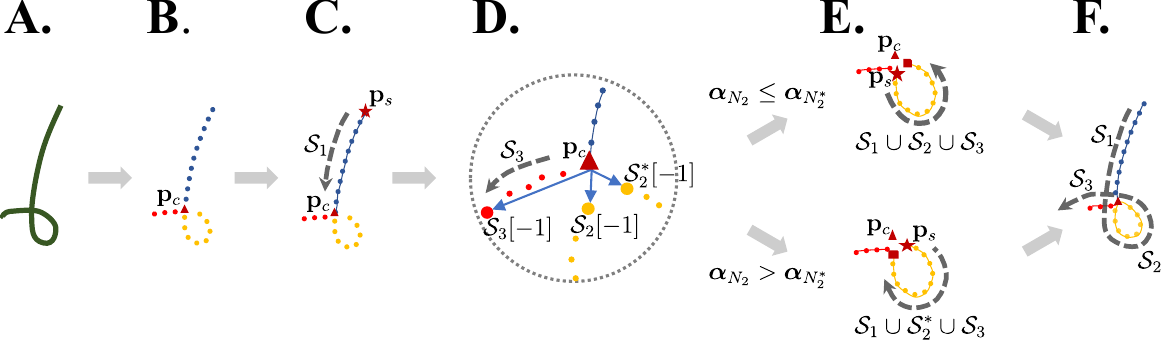}
\caption{Illustration of the sorting algorithm steps for off-plane morphing where the distance between the self-intersection point and tendril tip is not negligible. $\mathbf{A}.$ Example of tendril configuration. $\mathbf{B}.$ Unordered points of the skeleton. $\mathbf{C}.$ The ordering starts from $\mathbf{p}_{s}$ until reaching the point of intersection $\mathbf{p}_{c}$ that is then removed. The first set of ordered points is generated ($\mathcal{S}_1$). The remaining points are grouped according to their connectivity and sorted, obtaining three sets: one containing the tip ($\mathcal{S}_3$) and two on the ring having reversed order ($\mathcal{S}_2$, $\mathcal{S}^*_2$).     $\mathbf{D}.$ Each set is labeled according to the distance between $\mathbf{p}_{c}$ and the last point in the set. The one with the biggest distance (i.e., $\mathcal{S}_3$) is removed. $\mathbf{E}.$ The correct set between $\mathcal{S}_2$ and $\mathcal{S}^*_2$ is selected by considering the angles between the lines pairs $\langle\mathbf{L}_{N_1}, \mathbf{L}_{N_2} \rangle$ and $\langle\mathbf{L}_{N_1}, \mathbf{L}_{N^*_2} \rangle$. $\mathbf{F}.$ Final ordered set of points. }
\label{fig:cross-intersectionB}
\end{figure}

If the distance between the tip and the intersection is not negligible (Figure~\ref{fig:cross-intersectionB}), the algorithm divides the set of points into three subsets of points. Then, it proceeds with the automatic sorting as in the previous case. Starting from $\mathcal{S}_1$, it selects all the points from $\mathbf{p}_s$ up to the junction point, $\mathbf{p}_c$. Here, it computes the Euclidean distance from $\mathbf{p}_c$ to the last point of each set $\mathcal{S}_2$, $\mathcal{S}^*_2$, and $\mathcal{S}_3$. $\mathcal{S}_3$ contains all the points from the tip of the tendril to $\mathbf{p}_c$ (highlighted in red in Figure~\ref{fig:cross-intersectionB}B), and it is identified as such because the distance of the last point is the largest among all:

\begin{equation}
\displaystyle
	\begin{split}
 		& \lVert \mathbf{p}_c - \mathcal{S}_3[-1] \rVert > \lVert \mathbf{p}_c - \mathcal{S}_2[-1] \rVert  \\
 		& \lVert \mathbf{p}_c - \mathcal{S}_3[-1] \rVert > \lVert \mathbf{p}_c - \mathcal{S}^*_2[-1] \rVert 
	\end{split}
\end{equation}

with $[-1]$ pointing to the last element in a list. When all the sets of points are labeled and ordered, it is possible to merge the sets by evaluating at the angle between the lines $\mathbf{L}_{N_1}$, $\mathbf{L}_{N_2}$ and $\mathbf{L}_{N^*_2}$ as in the previous case: 
\begin{equation}
\displaystyle
	\begin{split}
   	 	\mathcal{S} = \left\{
    			\begin{split}
         			& \mathcal{S}_1 \cup \mathcal{S}_2 \cup \mathcal{S}_3, \quad if ~ \boldsymbol{\alpha}_{N_{2}} \leq \boldsymbol{\alpha}_{N^*_2} \\
         			& \mathcal{S}_1 \cup \mathcal{S}^*_2 \cup \mathcal{S}_3, \quad if ~ \boldsymbol{\alpha}_{N_2} > \boldsymbol{\alpha}_{N^*_2}.
   		 	\end{split}
    		\right.
	\end{split}
\end{equation}

\subsection{Mapping Points Between Views}
\label{sec:point_correspondence}

\begin{figure}
\centering
\includegraphics[width=\textwidth]{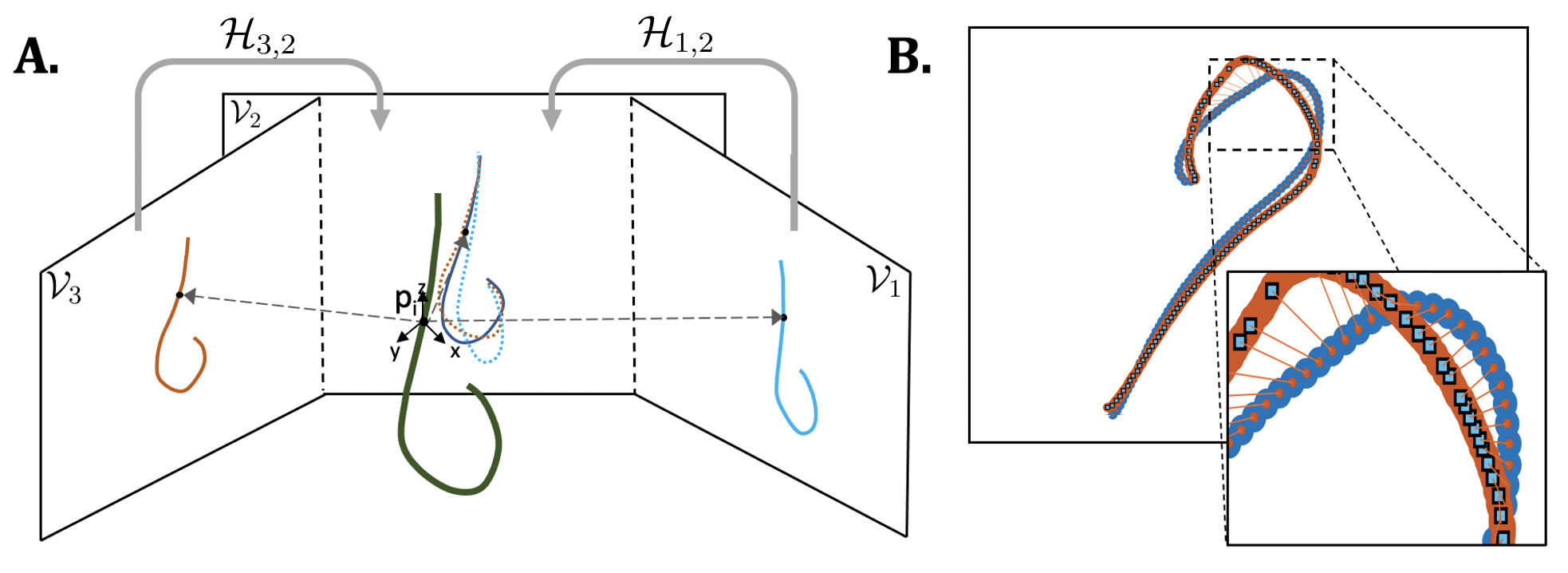}
\caption{The process of mapping the points between the views. $\mathbf{A}.$ Homographic transformation between the views to position all the points with respect to a single reference frame. $\mathbf{B}.$ Application of the Fr\'{e}chet distance to map the extracted set of points. When a set of correspondence has been found, the triangulation of the points can be applied to retrieve the three-dimensional shape of the curve.}
\label{fig:frechet}
\end{figure}

Once $\mathcal{S}$ is obtained in each $\mathcal{V}_1$, $\mathcal{V}_2$ and $\mathcal{V}_3$, we need to define a point-to-point correspondence between views. The first step involves a homographic projection between the views, taking $\mathcal{V}_2$ as a reference. The ArUco marker has been used as a reference in the computation of the homography matrix $\mathcal{H}_{1,2}$ and $\mathcal{H}_{3,2}$. As a result, we obtain a projection of the points in $\mathcal{V}_1$ and $\mathcal{V}_3$ onto $\mathcal{V}_2$ (Figure~\ref{fig:frechet}A), generating the point-to-point correspondence between views using the same reference. Given two curves $\mathcal{S}_p$ and $\mathcal{S}_q$, in a metric space composed of order-preserving sets of points, it is possible to find the correspondence between the curves using the Fr\'{e}chet distance \cite{Eiter_1994}. It measures the similarity between curves by considering the location and ordering of the points along them.

Let $\mathcal{S}_p = \left\lbrace u_1, . . . , u_p \right\rbrace$ and $\mathcal{S}_q = \left\lbrace v_1, . . . , v_q \right\rbrace$ be the corresponding sequences, we want to find the mapping between points:

\begin{equation}
	\left(u_{a_{1}}, v_{b_{1}}\right),\left(u_{a_{2}}, v_{b_{2}}\right), \ldots,\left(u_{a_{m}}, v_{b_{m}}\right)
\end{equation}

where $a_1=1$, $b_1=1$ and $a_m=p$, $b_m=q$, that minimizes the disparity for 3D triangulation-based reconstruction between the points.
The mapping also has to respect the order of the points by $a_{i+1} =a_{i} + 1$ and $b_{i+1} =b_{i} + 1$. Considering that the sets of points have been sorted, we define the discrete Fr\'{e}chet distance, denoted as $\delta_{F}\left(\mathcal{S}_p, \mathcal{S}_q\right)$ by:\\
\begin{equation}
	\begin{split}
   		\delta_{F}\left(\mathcal{S}_p, \mathcal{S}_q\right) &= \inf_{a_i, b_i} \max_{i=1, \ldots, m} \left\lbrace \rm{dis}\left(u_{a_{i}}, v_{b_{i}}\right) \right\rbrace \\
	\end{split}
\end{equation}
where $\rm{dis}$ is the Euclidean distance between one pair of points, and $\rm{\inf}$ is the infimum over all parameterization of $a_i$ and $b_i$. Before applying the Fr\'{e}chet distance, we uniformly sample $k~(k=200)$ points from each view for consistency during processing. After the mapping, the sampled points are back-projected in their original view using the same homographic transformation. Lastly, through triangulation, we project the 2D points in the 3D space to obtain the discrete 3D points:

\begin{equation}
    \mathcal{C} = \left\lbrace \mathbf{c}_1, \mathbf{c}_2, \cdots, \mathbf{c}_k \right\rbrace.
\end{equation}

\subsection{Discrete Frenet-Serrat Point Fitting}
To describe the shape of the tendril from the reconstructed 3D points, we adopt a 3D clothoid curve fitting with piece-wise approximation. We 
extended our work from 2D \cite{Fan_2020_2} to 3D by introducing information about the torsion.

Following the definition in the Frenet-Serret frame as in \cite{Harary_2010}, a 3D curve is defined by:
\begin{equation} 
	\label{eq:frenet_frame}
	\begin{split}
   		 \mathcal{C}(s) = \int_0^s \mathbf{T}\left(u\right) \mathrm{d}u + \mathbf{c}_0 = \int_0^s \left[ \int_0^t \frac{\mathrm{d} \mathbf{T}\left(u\right)}{\mathrm{d}u} \mathrm{d}u + \mathbf{T}_0 \right] \mathrm{d}t + \mathbf{c}_0
	\end{split}
\end{equation}

where the Frenet–Serret frame vectors are given by: 

\begin{equation} 
	\begin{split}
    		& \frac{\mathrm{d} \mathbf{T}\left(s\right)}{\mathrm{d}s} = \kappa(s) \mathbf{N}(s) \\
    		& \frac{\mathrm{d} \mathbf{N}\left(s\right)}{\mathrm{d}s} = - \kappa(s) \mathbf{T}(s) + \tau(s) \mathbf{B}(s) \\
    		& \frac{\mathrm{d} \mathbf{B}\left(s\right)}{\mathrm{d}s} = - \tau(s) \mathbf{N}(s)
	\end{split}
\end{equation}

and $\kappa(s)$ and $\tau(s)$ are respectively curvature and torsion, having a linear relationship with curve length:

\begin{equation}
	\label{eq:euler_linearity}
	\begin{split}
    		& \kappa(s) = \kappa_0' s + \kappa_0\\
    		& \tau(s) = \tau_0' s + \tau_0
	\end{split}
\end{equation}

In the case of a discrete set of points, as already investigated in \cite{Shuangwei_2011, Lim_2017}, it is possible to describe the Frenet-Serret frame as follows. For a given point $\mathbf{c}_i$ in the set of 3D points $\mathcal{C}$, we can define a sequence of the recursively discrete frame as $\mathcal{F}_i = \left(\mathbf{T}_i, \mathbf{N}_i, \mathbf{B}_i \right)$. The unit tangent vector is given:
\begin{equation} 
	\begin{split}
    		& \mathbf{T}_i = \frac{\mathbf{c}_{i+1}-\mathbf{c}_{i}}{| \mathbf{c}_{i+1}-\mathbf{c}_{i} |}
	\end{split}
\end{equation}

Then, the bi-normal and normal vectors can be defined by:

\begin{equation} 
	\begin{split}
    		& \mathbf{N}_i = \frac{\mathbf{T}_{i} - \mathbf{T}_{i-1}}{| \mathbf{T}_{i} - \mathbf{T}_{i-1} |}\\
    		& \mathbf{B}_i = \mathbf{T}_{i} \times \mathbf{N}_{i}
	\end{split}
\end{equation}

It should be noticed that there will be no discrete Frenet frames defined for the first and the last point due to the dependence of calculation requiring at least three neighboring points. From the above equations, the discrete curvature and torsion can be derived as:

\begin{equation}
	\begin{split}
    		& \kappa_i = \left\lVert \mathbf{T}_{i} - \mathbf{T}_{i-1} \right\rVert \\
    		& \tau_i = \frac{\mathbf{B}_{i} - \mathbf{B}_{i-1} }{|\mathbf{N}_{i} |}
	\end{split}
\end{equation}

\subsection{Automatic Piece-Wise Fitting of Curvature \& Torsion}
To guarantee the curvature-torsion continuity, we linearly fit the two quantities independently while preserving the linearity of the clothoid curve, as in Equation (\ref{eq:euler_linearity}). As in our previous implementation \cite{Fan_2020_2}, we used dynamic programming to minimize the number of line segments and the fitting error. The objective functions of our optimization problem are:

\begin{equation}
	\label{eq:costfunction}
	\begin{split}
    		& \mathcal{J}_{\boldsymbol{\kappa}}\left(\mathbf{c}_a, \mathbf{c}_b\right) =\min_{a<m<b} \left\lbrace \mathcal{J}_{\boldsymbol{\kappa}} \left(\mathbf{c}_a, \mathbf{c}_m\right) + \mathcal{J}_{\boldsymbol{\kappa}}\left(\mathbf{c}_m, \mathbf{c}_b\right), \epsilon_{\boldsymbol{\kappa}} + \mathcal{E}_{\boldsymbol{\kappa}}^{fit}\left(\mathbf{c}_a, \mathbf{c}_b\right) \right\rbrace \\
    		& \mathcal{J}_{\boldsymbol{\tau}}\left(\mathbf{c}_e, \mathbf{c}_f\right) =\min_{e<n<f} \left\lbrace \mathcal{J}_{\boldsymbol{\tau}} \left(\mathbf{c}_e, \mathbf{c}_n\right) + \mathcal{J}_{\boldsymbol{\tau}}\left(\mathbf{c}_n, \mathbf{c}_f\right), \epsilon_{\boldsymbol{\tau}} + \mathcal{E}_{\boldsymbol{\tau}}^{fit}\left(\mathbf{c}_e, \mathbf{c}_f\right) \right\rbrace
	\end{split}
\end{equation}

where $\mathcal{J}_{\left\lbrace \boldsymbol{\kappa}\right\rbrace}$ and $\mathcal{J}_{\left\lbrace\boldsymbol{\tau} \right\rbrace}$ are the costs of fitting error for curvature, $\boldsymbol{\kappa}$, and torsion,$\boldsymbol{\tau}$ between two points.

On the right hand of the equation,  $ \epsilon_{\left\lbrace \boldsymbol{\kappa} \right\rbrace}$ and $ \epsilon_{\left\lbrace \boldsymbol{\tau} \right\rbrace}$ are the constant penalty terms for adding a new segment. $\mathcal{E}_{\left\lbrace \boldsymbol{\kappa} \right\rbrace}^{fit}$ and $\mathcal{E}_{\left\lbrace \boldsymbol{\tau} \right\rbrace}^{fit}$ are the least-square fitting errors for linear regression of line segments, that is:

\begin{equation} 
	\label{eq:fit_line_error}
	\begin{split}
    		& \mathcal{E}_{\boldsymbol{\kappa} }^{fit}\left(\mathbf{c}_a, \mathbf{c}_b\right) = \min_{\alpha_{\boldsymbol{\kappa}}, \beta_{\boldsymbol{\kappa}}} \left(\sum_{t=a}^b | \alpha_{\boldsymbol{\kappa}} \cdot s(\mathbf{c}_t) + \beta_{\boldsymbol{\kappa}} - \kappa_t| \right) \\
    		& \mathcal{E}_{\boldsymbol{\tau} }^{fit}\left(\mathbf{c}_e, \mathbf{c}_f\right) = \min_{\alpha_{\boldsymbol{\tau}}, \beta_{\boldsymbol{\tau}}} \left(\sum_{t=e}^f | \alpha_{\boldsymbol{\tau}} \cdot s(\mathbf{c}_t) + \beta_{\boldsymbol{\tau}} - \tau_t| \right)
	\end{split}
\end{equation}

with $\alpha_{\left\lbrace \boldsymbol{\kappa} \right\rbrace}$, $\alpha_{\left\lbrace \boldsymbol{\tau} \right\rbrace}$, $\beta_{\left\lbrace \boldsymbol{\kappa}\right\rbrace}$, and $\beta_{\left\lbrace\boldsymbol{\tau} \right\rbrace}$ being the slopes and intercepts of the least-square fitting line segments. The variables $s(\mathbf{c}_t)$, $\kappa_t$ and $\tau_t$ are the arc-length, curvature and torsion at point $\mathbf{c}_t$, respectively.

Therefore, we can find the optimal automatic segmentation for curvature and torsion with $M$ and $N$ segments, respectively. Ideally, we expect the curvature/torsion continuity for neighboring pairs of segments. That is, for the intersection points between $i^{th}$ and ${(i+1)}^{th}$ segment at $\mathbf{c}_{\boldsymbol{\kappa},i+1}$ and $\mathbf{c}_{\boldsymbol{\tau},i+1}$, the curvature and torsion at the end-point of $i^{th}$ segment should be matched with the start-point of the $(i+1)^{th}$ segment as:

\begin{equation} 
	\label{eq:fit_line_segment_para}
	\begin{split}
    		& \alpha_{\boldsymbol{\kappa},i} \cdot s(\mathbf{c}_{\boldsymbol{\kappa},i+1}) + \beta_{\boldsymbol{\kappa}, i} =  \alpha_{\boldsymbol{\kappa},i+1} \cdot s(\mathbf{c}_{\boldsymbol{\kappa},i+1}) + \beta_{\boldsymbol{\kappa}, i+1} \\
    		& \alpha_{\boldsymbol{\tau},i} \cdot s(\mathbf{c}_{\boldsymbol{\tau},i+1}) + \beta_{\boldsymbol{\tau}, i} =  \alpha_{\boldsymbol{\tau},i+1} \cdot s(\mathbf{c}_{\boldsymbol{\tau},i+1}) + \beta_{\boldsymbol{\tau}, i+1}
	\end{split}
\end{equation}

Even though the start-end point between adjacent segments has been connected, the above curvature/torsion continuity may be violated because intersection constraints in both curvature and torsion space are not considered. We refined the segmentation point list by calculating slopes and intercepts for each pair of adjacent segments to solve this interconnection problem. However, this approach may cause an overshooting problem -- i.e., the refined point lies outside the two boundary points for the corresponding adjacent segments. Thus, we proposed a compensation procedure that combines these adjacent segments if overshooting occurs.

\subsection{Three-dimensional Reconstruction}  

At this point, we have the fitted curvature and torsion at each curve length. The 3D curve reconstruction can be done by integrating the TNB frame as in Equation~(\ref{eq:frenet_frame}). According to \cite{Lim_2017}, we can get the piece-wise integration using $4_{th}$ order Runge-Kutta method:

\begin{equation}
	\label{eq:runge_kutta_integration}
	\begin{split}
    		& \mathbf{T}_{i+1} = \left[1 + \frac{\boldsymbol{\kappa}^2_i \Delta s^2}{2} + \frac{(\boldsymbol{\kappa}^4_i +  \boldsymbol{\kappa}^2_i\boldsymbol{\tau}^2_i) \Delta s^4}{4} \right] \mathbf{T}_{i} + \left[\boldsymbol{\kappa}_i \Delta s - \frac{(\boldsymbol{\kappa}^3_i -  \boldsymbol{\kappa}_i\boldsymbol{\tau}^2_i) \Delta s^3}{6} \right] \mathbf{N}_{i} \\
    		&  ~~~~~~~~~ + \left[\frac{\boldsymbol{\kappa}_i\boldsymbol{\tau}_i \Delta s^2}{2} - \frac{(\boldsymbol{\kappa}^3_i\boldsymbol{\tau}_i +  \boldsymbol{\kappa}_i\boldsymbol{\tau}^3_i) \Delta s^4}{24} \right] \mathbf{B}_{i} \\
    		& \mathbf{N}_{i+1} = \left[ - \boldsymbol{\kappa}_i \Delta s + \frac{(\boldsymbol{\kappa}_i\boldsymbol{\tau}^2_i +  \boldsymbol{\kappa}^3_i) \Delta s^3}{6} \right] \mathbf{T}_{i} + \left[ 1 - \frac{(\boldsymbol{\kappa}^2_i+\boldsymbol{\tau}^2_i) \Delta s^2}{2} + \frac{(\boldsymbol{\kappa}^2_i+\boldsymbol{\tau}^2_i)^2 \Delta s^4}{24} \right] \mathbf{N}_{i} \\
    		&  ~~~~~~~~~ + \left[\boldsymbol{\tau}_i \Delta s - \frac{(\boldsymbol{\kappa}^2_i\boldsymbol{\tau}_i +  \boldsymbol{\tau}^3_i) \Delta s^3}{6} \right] \mathbf{B}_{i} \\
    		& \mathbf{B}_{i+1} = \left[ \frac{\boldsymbol{\kappa}_i \boldsymbol{\tau}_i \Delta s^2}{2} - \frac{(\boldsymbol{\kappa}_i\boldsymbol{\tau}^3_i +  \boldsymbol{\kappa}^3_i \boldsymbol{\tau}_i) \Delta s^4}{24} \right] \mathbf{T}_{i} 
    + \left[ - \boldsymbol{\tau} \Delta s + \frac{(\boldsymbol{\kappa}^2_i\boldsymbol{\tau}_i+\boldsymbol{\tau}^3_i) \Delta s^3}{6} \right] \mathbf{N}_{i} \\
    		&  ~~~~~~~~~ + \left[1 - \frac{\boldsymbol{\tau}^2_i \Delta s^2}{2} - \frac{(\boldsymbol{\kappa}^2_i\boldsymbol{\tau}^2_i +  \boldsymbol{\tau}^4_i) \Delta s^4}{24} \right] \mathbf{B}_{i} \\
	\end{split}
\end{equation}

The approximation of the integrated curve point can be achieved by the trapezoidal rule as follows:

\begin{equation}
	\label{eq:integrated_curve_points}
	\begin{split}
    		& \mathcal{P}_{i+1} = \mathcal{P}_{i} + \frac{\Delta s}{2} \left( \mathbf{T}_{i+1} + \mathbf{T}_{i}\right)
	\end{split}
\end{equation}

where the initial point is $\mathbf{p}_{0} = \mathbf{c}_{0}$ from the original ordered discrete points set.

The integrated TNB frame depends on the initial TNB vectors (i.e., $\mathbf{T}_0, \mathbf{N}_0, \mathbf{B}_0$). Thus, its accuracy may be affected by a noisy initialization. To improve the global reconstruction performance, we can perform an optimal rigid transformation according to shape matching theory \cite{Bouaziz_2016}.

If $\mathcal{C} = \left\lbrace \mathbf{c}_1, \mathbf{c}_2, \cdots, \mathbf{c}_t \right\rbrace$ is the set of ordered discrete points from multi-view reconstruction, and $\mathcal{P} = \left\lbrace \mathbf{p}_1, \mathbf{p}_2, \cdots, \mathbf{p}_t \right\rbrace$ are the points obtained from the shape modeling using 3D PWC, finding the optimal rotation matrix $\mathbf{R}$ and the translation vector $[\mathbf{t}, \mathbf{t}_0]$ to register the two curves is a least-square optimization problem:

\begin{equation}
	\begin{split}
        		\left(\mathbf{R}, \mathbf{t}\right) = \arg\min_{\mathbf{R}, \mathbf{t}} \sum_{i=1}^t \left\lVert \mathbf{R}\cdot(\mathbf{p}_i-\mathbf{t}_0) + \mathbf{t} - \mathbf{c}_i \right\rVert_2^2
    	\end{split}
\end{equation}

that can be solved using singular value decomposition (SVD)\cite{Bouaziz_2016}.

\section{Results}\label{section:Results}
To validate our approach, we deliberately selected a dataset in which stimulation was applied to the 3rd section of a tendril (Figure~\ref{fig:expri_setup}D). This dataset was chosen because its morphological evolution exhibited partially similar patterns observed when other sections (i.e., the 1st, 2nd and 4th sections) were stimulated. As a result, the morphological changes in these other sections could be considered subsets of the dataset used for verification. To avoid redundancy, we focus solely on presenting results from this dataset. Furthermore, 
this dataset effectively captures the variability in response to stimulation, as different sections exhibited varying degrees of bending and intersection cases, making it a representative choice for validating our approach. For analysis, we selected three instances representative of the particular tendril's morphing phase: the beginning (at $1^{st}$ frame), half response (at $100^{th}$ frame), and final shape response (at $200^{th}$ frame) after stimulation. 

\begin{figure}[htb]
\centering 
\includegraphics[width=\textwidth]{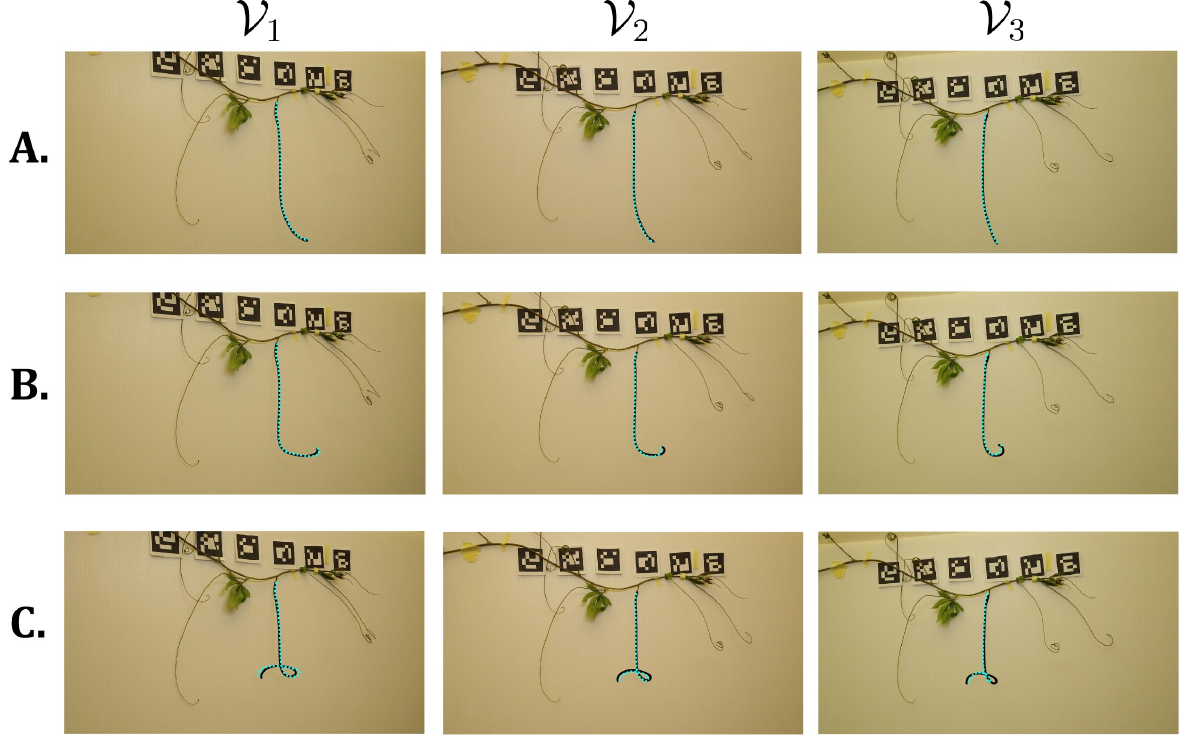} 
\caption[Shape comparison of reconstruction accuracy between tendril skeletons and projections of the stereo reconstructed tendril]{The shape comparison of extracted tendril skeletons (highlighted in black) and the stereo reconstructed projections (highlighted in cyan) in corresponding views, $\mathcal{V}_1$, $\mathcal{V}_2$ and $\mathcal{V}_3$. \textbf{A}, \textbf{B} and \textbf{C} are the results of the $1^{st}$ frame, the $100^{th}$ frame and the $200^{th}$ frame, shown by row.}	\label{fig:ProjectionError_Skeleton_Stereo}
\end{figure}

\begin{table}[hb!]
    \centering
    \resizebox{\textwidth}{25mm}{
    \begin{NiceTabular}{|c|c|rrr|rrr|}
    \CodeBefore
      \rowcolor[gray]{0.95}{1,2}
      \rowcolors{3}{}{}
    \Body
    \Hline
      \Block{2-1}{Frame} &  \Block{1-1}{}& \Block{1-3}{Mean Error (pixel)}& & & \Block{1-3}{Standard Deviation (pixel)}\\
    \cline{2-8}
        & Views  & Base Seg & Tip Seg  & Entire & Base Seg  & Tip Seg & Entire \\
    \Hline
      {}  & $\mathcal{V}_1$  & 2.9020  & 3.0643  & 3.4951  & 0.4448  & 0.4390  & 0.8043\\
      1   & $\mathcal{V}_2$  & 3.7757  & 9.5178  & 4.9690  & 0.6733  & 1.0568  & 2.9742\\
      {}  & $\mathcal{V}_3$  & 6.8500  & 7.0944  & 8.0666  & 1.3554  & 1.9023  & 1.9148\\ \hline
      {}  & $\mathcal{V}_1$  & 3.2556  & 15.0116 & 7.9629  & 0.6560  & 3.3446  & 4.8738\\
      100 & $\mathcal{V}_2$  & 6.8103  & 16.3470 & 8.6539  & 2.0423  & 2.8152  & 5.2214\\
      {}  & $\mathcal{V}_3$  & 9.1724  & 22.2579 & 17.6280 & 1.3888  & 6.7207 & 7.0410\\ \hline
      {}  & $\mathcal{V}_1$  & 5.7096  & 7.1049  & 13.0010 & 2.0893  & 5.8880 & 8.6477\\
      200 & $\mathcal{V}_2$  & 6.1356  & 10.3278 & 10.9350 & 2.5911  & 1.5974 & 5.6904\\
      {}  & $\mathcal{V}_3$  & 13.1250 & 10.4357 & 22.4540 & 3.2229  & 6.2604 & 13.0686\\
    \Hline
    \end{NiceTabular}}
    \caption{Mean errors and standard deviations of the stereo reconstruction for the entire tendril, only the base segment, and only for the tip segment. Results refer to three morphing phases: at the $1^{st}$, $100^{th}$, and $200^{th}$ frame.}
    \label{Tab:ProjectionError_Skeleton_Stereo}
\end{table}

\begin{figure}[]
    \centering  \includegraphics[width=0.9\textwidth]{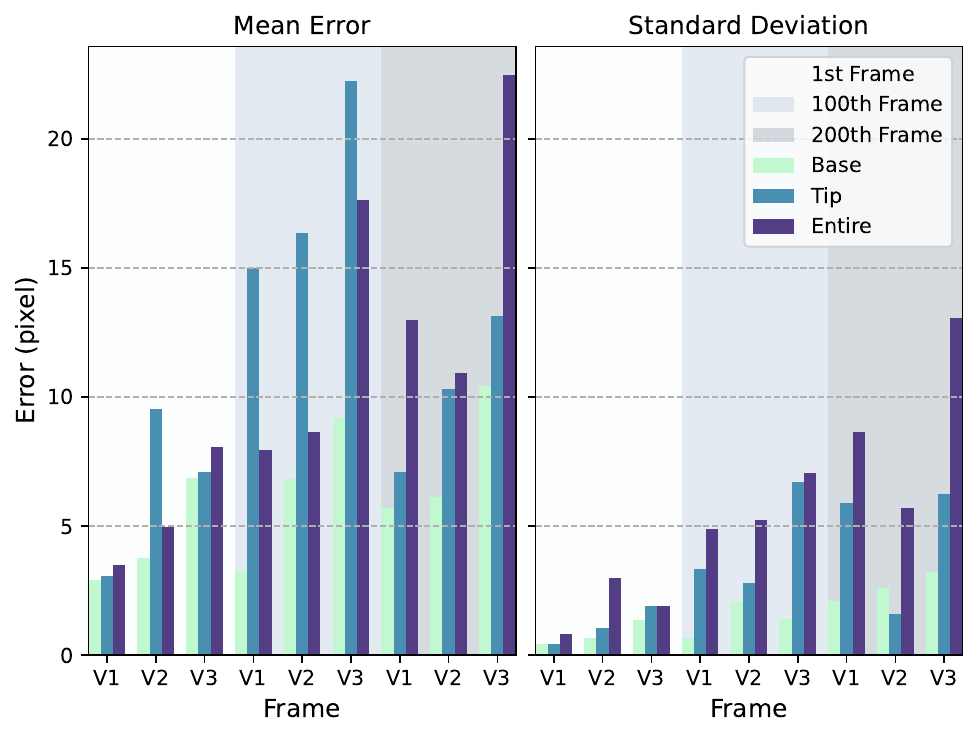}
    \caption[bar plot of mean errors and standard deviations of the stereo reconstruction]{The bar plots of mean errors and standard deviations of the stereo reconstruction for the entire tendril, only the base segment, and only the tip segment. Results refer to three morphing phases: at the $1^{st}$, $100^{th}$, and $200^{th}$ frame, coloring frame regions in white, light blue, and gray shades.}
    \label{Fig:BarPlot_3dRecons}
\end{figure}

\subsection{Accuracy of Stereo Reconstruction}
To verify the accuracy of our stereo reconstruction method, we projected the reconstructed 3D tendrils points onto the 2D images on each perspective (i.e., $\mathcal{V}_1$, $\mathcal{V}_2$ and $\mathcal{V}_3$) and compared them with the extracted 2D skeletons. In  Figure~\ref{fig:ProjectionError_Skeleton_Stereo}), we can observe a good match between the re-projections (highlighted in cyan) in the three morphing phases and the extracted skeleton (highlighted in black) in each view, with better accuracy in $\mathcal{V}_1$ and $\mathcal{V}_2$. 
This is because the bundle adjustment optimization process involved in the reconstruction uses $\mathcal{V}_2$ as a reference view, and $\mathcal{V}_1$ manifests closer similarity to $\mathcal{V}_2$ with respect to $\mathcal{V}_3$.
Conversely, the $\mathcal{V}_3$ perspective contributes to the shape optimization of the 3D model with respect to depth estimation.

To quantify the accuracy, we compared the pixel count of the extracted skeleton points with the projections of the reconstructed tendril points onto three image planes. This comparison focused on three sections: the base, the tip, and the entire tendril. The tip section was specifically analyzed because it undergoes the most significant morphological changes, while the base section exhibits the least deformation. These comparisons allow us to assess how morphological differences affect reconstruction accuracy.

The mean errors and standard deviations for each section were calculated by taking the absolute difference between the skeletonized points and the projection of reconstructed points, averaging over all the points in the section, and calculating their standard deviations. The base and tip sections were each defined to be one-quarter of the whole tendril length. 

Due to the low mobility and morphological changes of the base section of the tendril, the reconstruction of this part is more accurate than the other sections (lower mean error in Table~\ref{Tab:ProjectionError_Skeleton_Stereo} and Figure~\ref{Fig:BarPlot_3dRecons}). However, the mean error values over the tip and entire tendril remain within an acceptable range: always less than 23 pixels, which is quite small compared to the overall size of $2.0736 \times 10^6$ pixels. The standard deviation values of all sections are relatively small, indicating that the reconstructed tendril curve is consistent and reliable across frames. 

Furthermore, views $\mathcal{V}_1$ and $\mathcal{V}_2$ exhibit lower mean error and standard deviation compared to $\mathcal{V}_3$. This is because they serve as the primary sources of the 2D features used in reconstruction, maintaining stronger alignment with the reconstructed 3D points and resulting in superior re-projection accuracy for all sections and frames.

Besides, the errors become more significant as the tendril shape evolves from the initial morphology ($1^{st}$ frame) to the final shape ($200^{th}$ frame). This is because we use the first frame for tendril shape initialization, as the tendril morphs over time, the errors would add up and become pronounced.

\begin{figure}[]
    \centering 
        \includegraphics[width=\textwidth]{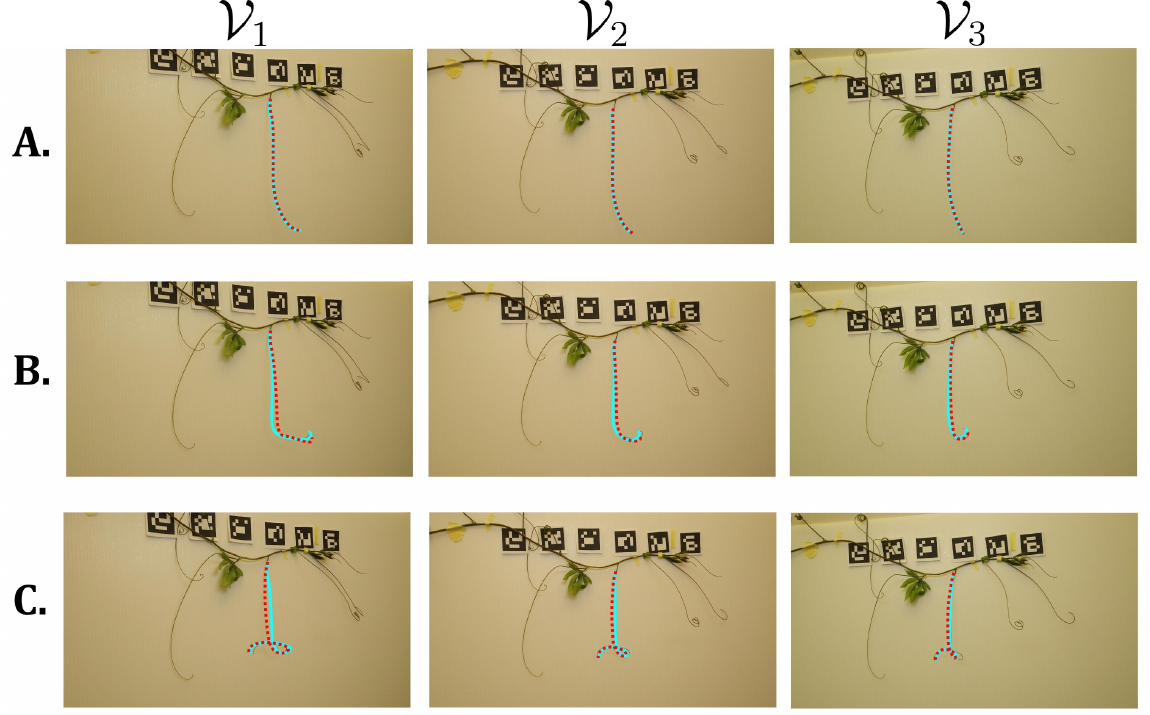}
    \caption[Shape comparison of fitting accuracy between projections of stereo reconstruction tendril and PWC fitted tendril]{The shape comparison of the stereo reconstruction projections (highlighted in cyan) and the PWC fitted projections (highlighted in red) in corresponding views, $\mathcal{V}_1$, $\mathcal{V}_2$ and $\mathcal{V}_3$. \textbf{A}, \textbf{B} and \textbf{C} are the results of the $1^{st}$ frame, the $100^{th}$ frame and the $200^{th}$ frame, shown by row.}
    \label{Fig:ProjectionError_PCD_Stereo}
\end{figure}

\begin{table}[]
    \centering
    \resizebox{\textwidth}{25mm}{
    \begin{NiceTabular}{|c|c|rrr|rrr|}
    \CodeBefore
      \rowcolor[gray]{0.95}{1,2}
      \rowcolors{3}{}{}
    \Body
    \Hline
      \Block{2-1}{Frame} &  \Block{1-1}{}& \Block{1-3}{Mean Error (pixel)}& & & \Block{1-3}{Standard Deviation (pixel)}\\
    \cline{2-8}
        & Views & Base Seg & Tip Seg & Entire & Base Seg  & Tip Seg & Entire\\
    \Hline
      {} & $\mathcal{V}_1$  & 3.7479  & 4.8391  & 6.2466  & 1.6500  & 1.5890  & 2.7127\\
      1  & $\mathcal{V}_2$  & 2.4242  & 6.3898  & 5.3262    & 1.0712  & 0.9608 & 2.0181\\
      {} & $\mathcal{V}_3$  & 2.1449  & 5.0599  & 4.1638  & 0.8736  & 1.5731  & 1.4418\\ \hline
      {} & $\mathcal{V}_1$  & 15.1840 & 10.3420 & 15.3940 & 5.9630  & 2.5095 & 5.2313 \\
      100& $\mathcal{V}_2$  & 8.1449  & 7.2112  & 9.4546  & 3.5505  & 2.1564  & 3.1676\\
      {} & $\mathcal{V}_3$  & 4.4073  & 4.5838  & 6.1265  & 2.1925  & 1.2073  & 2.3424\\ \hline
      {} & $\mathcal{V}_1$  & 15.2359 & 13.5972 & 16.2491  & 9.2382  & 1.3979 & 7.4025\\
      200& $\mathcal{V}_2$  & 8.7959  & 9.4189  & 10.5715  & 5.5375  & 0.9723 & 4.3685\\
      {} & $\mathcal{V}_3$  & 5.5857  & 5.4172  & 7.9385  & 3.7365  & 1.5812  & 3.6088\\
    \Hline
    \end{NiceTabular}}
    \caption{Mean errors and standard deviations between projections of reconstructed tendril points and piece-wise clothoid fitted points. Results relate to three morphing phases at the $1^{st}$, $100^{th}$, and $200^{th}$ frame.}
    \label{Tab:ProjectionError_PCC_Stereo}
\end{table}

\subsection{Accuracy of Clothoid Fitting}
We also used the re-projection approach and compared the projections onto the correspondent views to validate the accuracy of reconstructed points fitted by the PWC curves method. Figure \ref{Fig:ProjectionError_PCD_Stereo} illustrates that the stereo reconstruction projections (highlighted in cyan) and the PWC fitted projections (highlighted in red) have good fitting results for the whole shape in each corresponding view. Intuitively, the fitting results in the middle sections have a certain deviation.

\begin{figure}[]
    \centering 
        \includegraphics[width=0.9\textwidth]{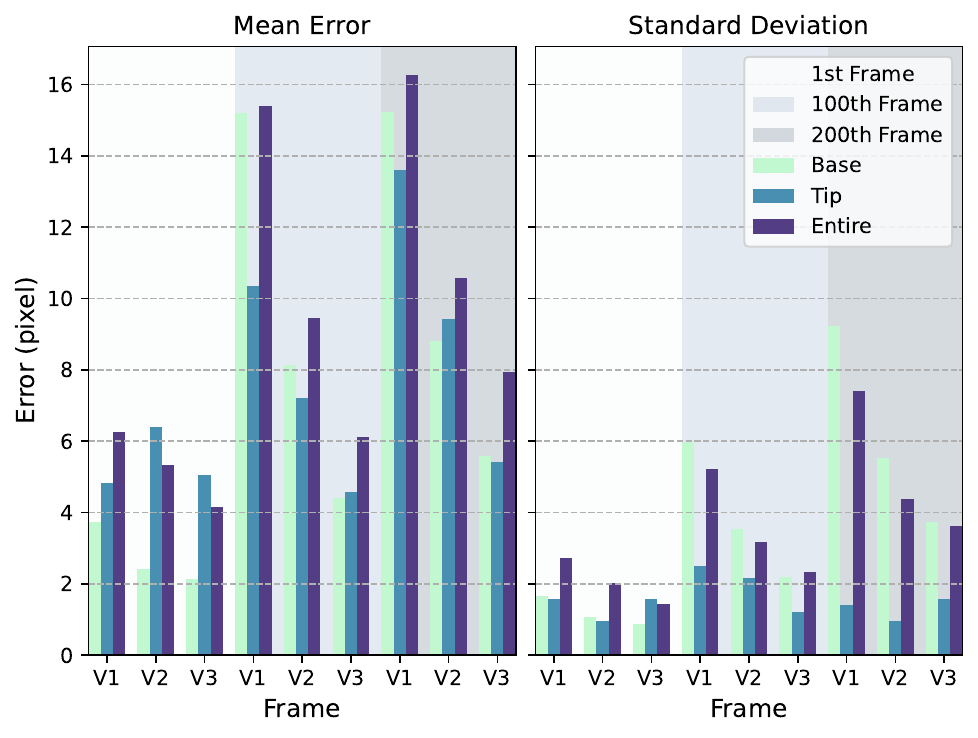}
    \caption[bar plot of mean errors and standard deviations of the piece-wise clothoid fitting]{The bar plots of mean errors and standard deviations between projections of reconstructed tendril points and piece-wise clothoid fitted points for the only base segment, the tip segment, and the entire tendril. Results refer to three morphing phases: at the $1^{st}$, $100^{th}$, and $200^{th}$ frame, coloring frame regions in white, light blue and gray shades.}
    \label{Fig:BarPlot_PCD}
\end{figure}

To quantify the accuracy of the PWC curve fitting model, we compared the image-pixel differences between the projections of the PWC curve fitting model and the original reconstructed tendril points. 
Table~\ref{Tab:ProjectionError_PCC_Stereo} shows the mean error and standard deviation for the three cases: the base, tip, and entire segments. They are visualized by bar plots in Figure~\ref{Fig:BarPlot_PCD} as well.
Because the fitting model is based on the 3D reconstruction output, the errors in the accuracy have a comparable behavior. This means that the errors overall become more significant with time, from the initial shape ($1^{st}$ frame) to the final shape ($200^{th}$ frame). However, differently from the reconstruction accuracy, we can observe the inaccuracy on the entire tendril, which tends to be larger than the tip sections for frames $> 100^{th}$, suggesting that the contribution for morphological evolution of the middle sections becomes more important here. All the standard deviation values are relatively small (less than 10 pixels), indicating that our clothoid fitting curves are reliable.

\subsection{Sensitivity of Fitting Parameters}
\begin{figure}[!htb]
	\centering 
    	\includegraphics[width=\linewidth]{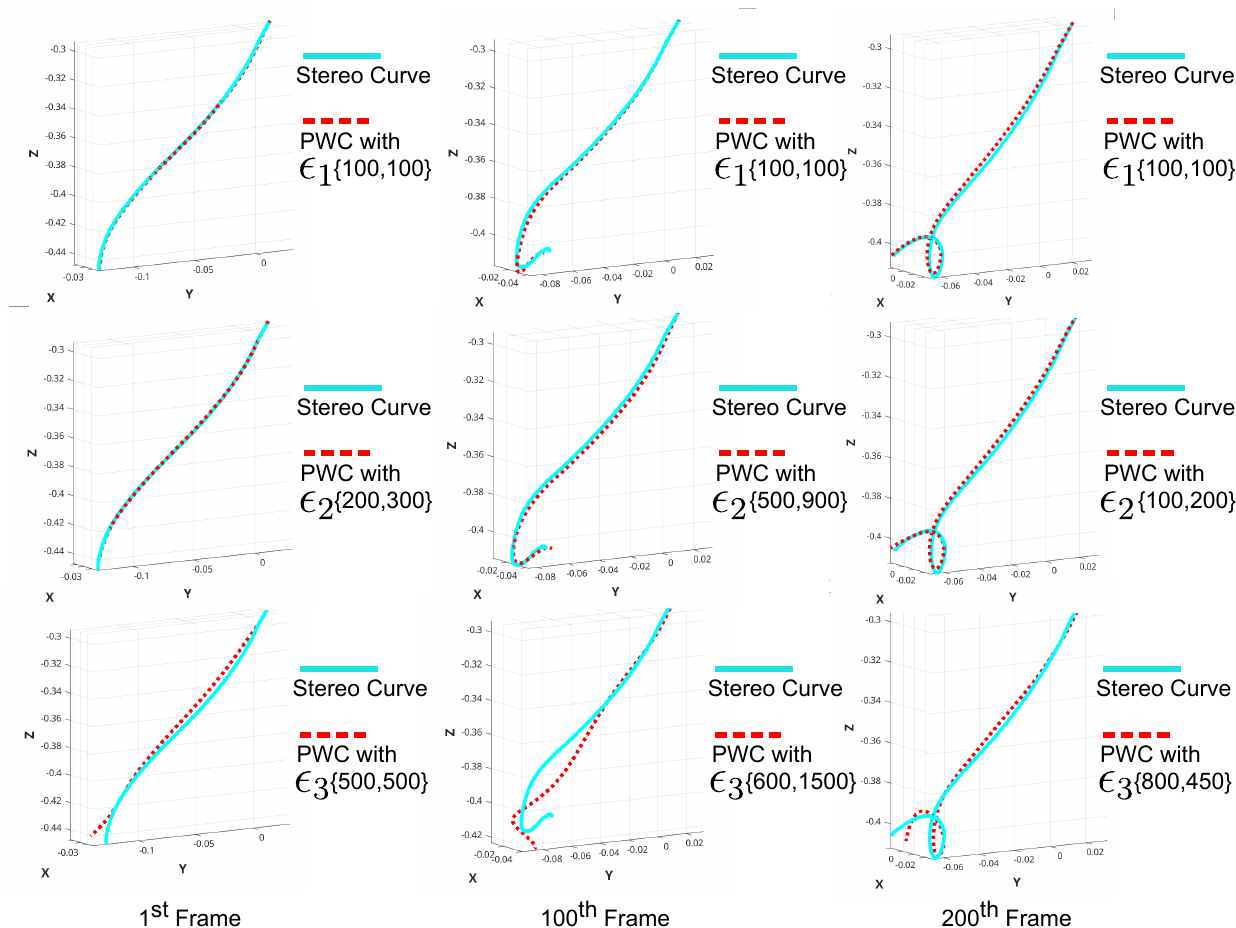} 
    	\caption[Piece-wise Clothoid fitting under different penalties for selected frames]{A fitting visualization of Table \ref{Tab:PCD_FittError}: three selected frames showing the stereo reconstructions in different morphological stages and fittings by Piece-Wise Clothoid curve under different penalties for curvature and torsion.}
    	\label{fig:PCD_fitting}
\end{figure}

The sensitivity of fitting parameters focuses on assessing how the fitting model responds to different numbers of segments, which is related to the linear piece-wise fitting of curvature and torsion. To evaluate this,
we set different, arbitrary chosen penalties ($\epsilon_{i\left\lbrace \boldsymbol{\kappa} \right\rbrace}$ and $\epsilon_{i\left\lbrace  \boldsymbol{\tau} \right\rbrace}$, where $i=1,2,3$) for the optimization problem in Equation  (\ref{eq:costfunction}) and evaluated the coefficient of determination ($R^2$) and the Sum of Squares Error (SSE) for PWC curve fitting.
The higher the penalties, the lower the number of the segments employed in the PWC reconstruction, potentially sacrificing accuracy to keep the fitting process simpler.
As anticipated, the performance declined as the number of segments decreased (as Table~\ref{Tab:PCD_FittError}). Additionally, at the $100^{th}$ frame, the tendril undergoes significant changes in both curvature and torsion that highly affect reconstruction goodness. A qualitative comparison is illustrated in Figure~\ref{fig:PCD_fitting}.

Although most cases achieve an $R^2$ value larger than $0.9$ (shown in Table~\ref{Tab:PCD_FittError}), exhibiting consistent performance across different parameter sets, the SSE shows variable results. Therefore, an automated selection of penalty values enables determining the optimal number of segments, ensuring an optimal fit for each frame by minimizing SSE and maximizing $R^2$. Given that penalty values are sampled within a predefined range, the problem can be effectively addressed using a grid search strategy.

\begin{figure}[!htb]
	\centering 
        \includegraphics[width=\linewidth]{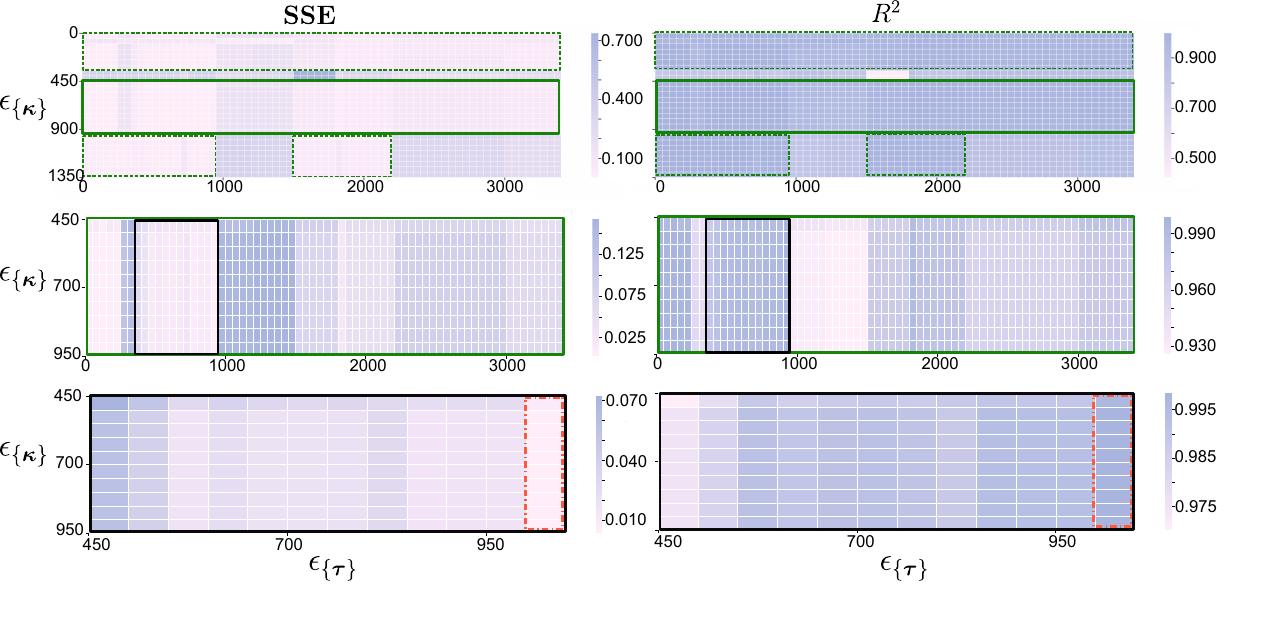} 
    	\caption[An Example of Grid Search Process for Selecting Penalty Values in PWC Fitting]{Illustration of the grid search process for selecting optimal penalty values in PWC fitting. The top row shows the initial search across a predefined range of penalty values for curvature and torsion. Based on the combined criteria of maximizing $R^2$ and minimizing SSE, four candidate regions (marked by the green rectangles) are identified. For illustrative purposes, one of these regions (highlighted by the green solid line) is selected for further refinement. The middle row indicates the next search step within this selected region. By applying the same evaluation criteria, a smaller candidate region (outlined by the black solid line) is chosen for continued analysis. This iterative grid search process continues, progressively narrowing the parameter space, until the optimal penalty values are determined based on the balance between $R^2$ and SSE. If the optimal values fall within regions satisfying $R^2$ $> 0.999$ and $SSE < 0.001$ (as shown in the bottom row indicated by the red dashed line), an arbitrary parameter pair from this range can be selected as the optimal penalty set of curvature and torsion.}
    	\label{Fig:GridSearch}
\end{figure}

Figure \ref{Fig:GridSearch} depicts the search process, taking the penalty selection for the $100^{th}$ frame as an example. We performed a grid search policy within a broad range, with the initial range of values defined as $\epsilon_{\left\lbrace \boldsymbol{\kappa} \right\rbrace} \in [0, 1350]$  and $\epsilon_{\left\lbrace \boldsymbol{\tau} \right\rbrace} \in [0, 3450]$, established through preliminary experiments to encompass the spectrum from minimal regularization to excessive smoothing. We then selected several regions of interest (e.g., the four regions within the green rectangles in the top row of Figure \ref{Fig:GridSearch}) based on combined criteria of maximizing $R^2$ and minimizing SSE, and repeated the grid search iteratively, progressively narrowing the parameter space until we found the appropriate penalties where fitting converged to $R^2>0.999$ and $SSE<0.001$. The computational cost of this process is approximately 200 seconds per frame on a workstation equipped with a 2.9 GHz quad-core Intel Core i7 processor, 16 GB RAM, and AMD Radeon Pro 560 GPU (4 GB VRAM), with convergence typically achieved within 3-4 refinement iterations. 
 
\begin{table}[]
    \centering
    \resizebox{\textwidth}{20mm}{
    \begin{NiceTabular}{|c|rrr|rr|rrr|rrr|}
    \CodeBefore
      \rowcolor[gray]{0.95}{1,2}
      \rowcolors{3}{}{}
    \Body
    \Hline
      \Block{2-1}{Frame} & \Block{1-3}{Penalty} &&& \Block{1-2}{Segs} && \Block{1-3}{$R^2$} &&& \Block{1-3}{SSE}\\
    \cline{2-12}
         & $i$ & $\epsilon_{i\left\lbrace \boldsymbol{\kappa} \right\rbrace}$ & $\epsilon_{i\left\lbrace \boldsymbol{\tau}\right\rbrace}$ & $\boldsymbol{\kappa}$ & $\boldsymbol{\tau}$ & Entire & Base Seg & Tip Seg  & Entire &Base Seg  &Tip Seg\\
    \Hline
      {}  & 1 & 100 & 100 & 6 & 136 & 0.9995 & 0.9977 & 0.9917 & 0.0014 & 0.0085 & 0.0006\\ 
      1   & 2 & 200 & 300 & 3 & 56  & 0.9993 & 0.9989 & 0.9989 & 0.0018 & 0.0040 & 0.0006\\ 
      {}  & 3 & 500 & 500 & 3 & 49  & 0.9717 & 0.6589 & 0.7749 & 0.0788 & 0.0127 & 0.0180\\
      \hline
      {}  & 1 & 100 & 100 & 13& 117 & 0.9963 & 0.9967 & 0.8540 & 0.0078 & 0.0001 & 0.0030 \\ 
      100 & 2 & 500 & 900 & 7 & 30  & 0.9983 & 0.9854 & 0.9645 & 0.0034 & 0.0005 & 0.0007\\ 
      {}  & 3 & 600 & 1500& 7 & 3   & 0.9699 & 0.9652 & 0.1882 & 0.0660 & 0.0011 & 0.0254\\
      \hline
      {}  & 1 & 100 & 100 & 18 & 112 & 0.9973 & 0.9826 & 0.9762 & 0.0045 & 0.0007 & 0.0005\\ 
      200 & 2 & 100 & 200 & 18 & 77  & 0.9977 & 0.9865 & 0.9801 & 0.0038 & 0.0005 & 0.0005\\ 
      {}  & 3 & 800 & 450 & 9  & 41  & 0.9850 & 0.9246 & 0.7727 & 0.0253 & 0.0031 & 0.0052\\
    \Hline
    \end{NiceTabular}}
    \caption{A summary of the results for the Piece-Wise Clothoid model fitting as a function of the penalty parameters. The results are relative to the tendril stimulated at \textbf{S3} and for the $1^{st}$, $100^{th}$, and $200^{th}$ frame. To easily compare the results, for every pair of parameters $\epsilon_{i\left\lbrace \boldsymbol{\kappa} \right\rbrace}$ and $\epsilon_{i\left\lbrace \boldsymbol{\tau} \right\rbrace}$, the number of segments needed for the approximation, the coefficient of determination ($R^2$) and the Sum of Squares Error (SSE) for Piece-Wise Clothoid Fitting are listed.}
    \label{Tab:PCD_FittError}
\end{table}

\subsection{Reconstruction From a Sequence of Frames}

\begin{figure}[tbp]
    \centering
    \includegraphics[width=\textwidth]{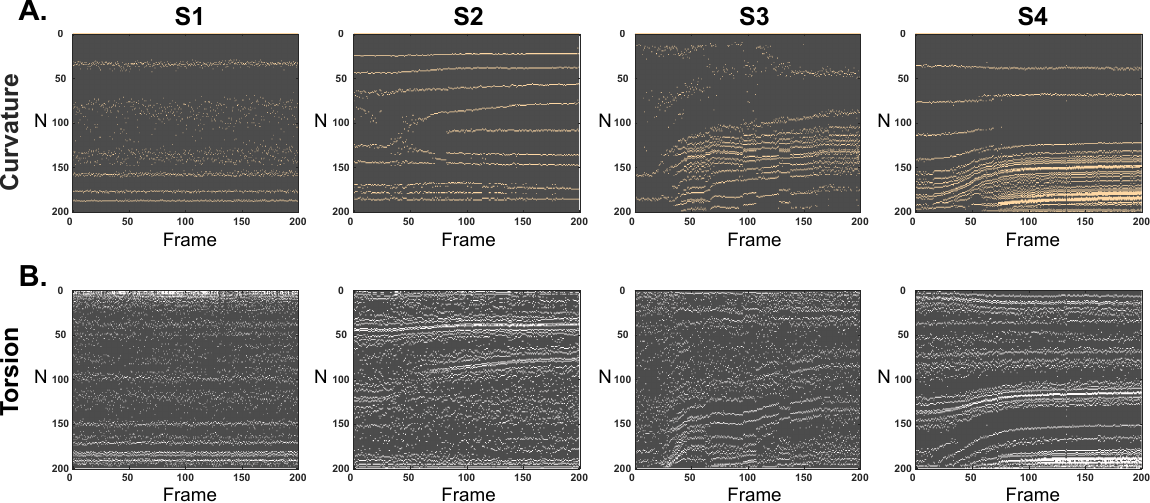}
    \caption{Comparison of the splitting point locations between segments in the linear fitting process for PWC curve fitting. Row \textbf{A}: shows the splitting points identified during curvature fitting.
    Row \textbf{B}: the splitting points identified during torsion fitting.
    For each column, the four different tendrils were stimulated in regions \textbf{S1}, \textbf{S2}, \textbf{S3}, \textbf{S4}, respectively.
    And we describe the reconstructed curve using down-sampled 200 points over a subset of the entire video sampled every $120~s$, reaching a number of frames of 200.}
    \label{fig:SegPtsPos}
\end{figure}

To analyze the morphological evolution of the tendrils over time, we must look at the sequence of frames acquired from the videos ($15$~min bring $\approx21,500$ frames). We fixed the number of points in the curve. All the curves have been equally down-sampled into 200 points after the 3D reconstruction, and the videos sampled every $120~s$ for a total of 200 frames.
To compare the effects of the applied stimuli, here we consider all the four cases, i.e., stimuli applied in region \textbf{S1}, \textbf{S2}, \textbf{S3}, \textbf{S4}, as introduced in Figure \ref{fig:expri_setup}D.

Figure \ref{fig:SegPtsPos} shows examples of how tendrils are split by the algorithm for each case in PWC curve fitting. The pixels index used to split into segments curvature and torsion are marked in a light color in the figure. 
From the graphs, we notice that the number of piece-wise segments is generally the same for all frames and that the index locations stabilize after a while (e.g., Figure \ref{fig:SegPtsPos}A, segment $\textbf{S2}$, from frame 80). In addition, the indexes are denser around the apical regions of the tendrils (pixels greater than $150$), especially for case \textbf{S3} and \textbf{S4}.
The number of segments to ascribe torsion
(figure \ref{fig:SegPtsPos}B) is even higher. Thus, we can tell that the torsion has a more nonlinear behavior along the entire length of the tendrils. From a mechanics point of view, the reconstructed curve shows non-constant twisting in space due to an additional out-of-plane torque. This pattern fits the requirements of tendrils to find support.

\begin{figure}
    \centering
    \includegraphics[width=\textwidth]{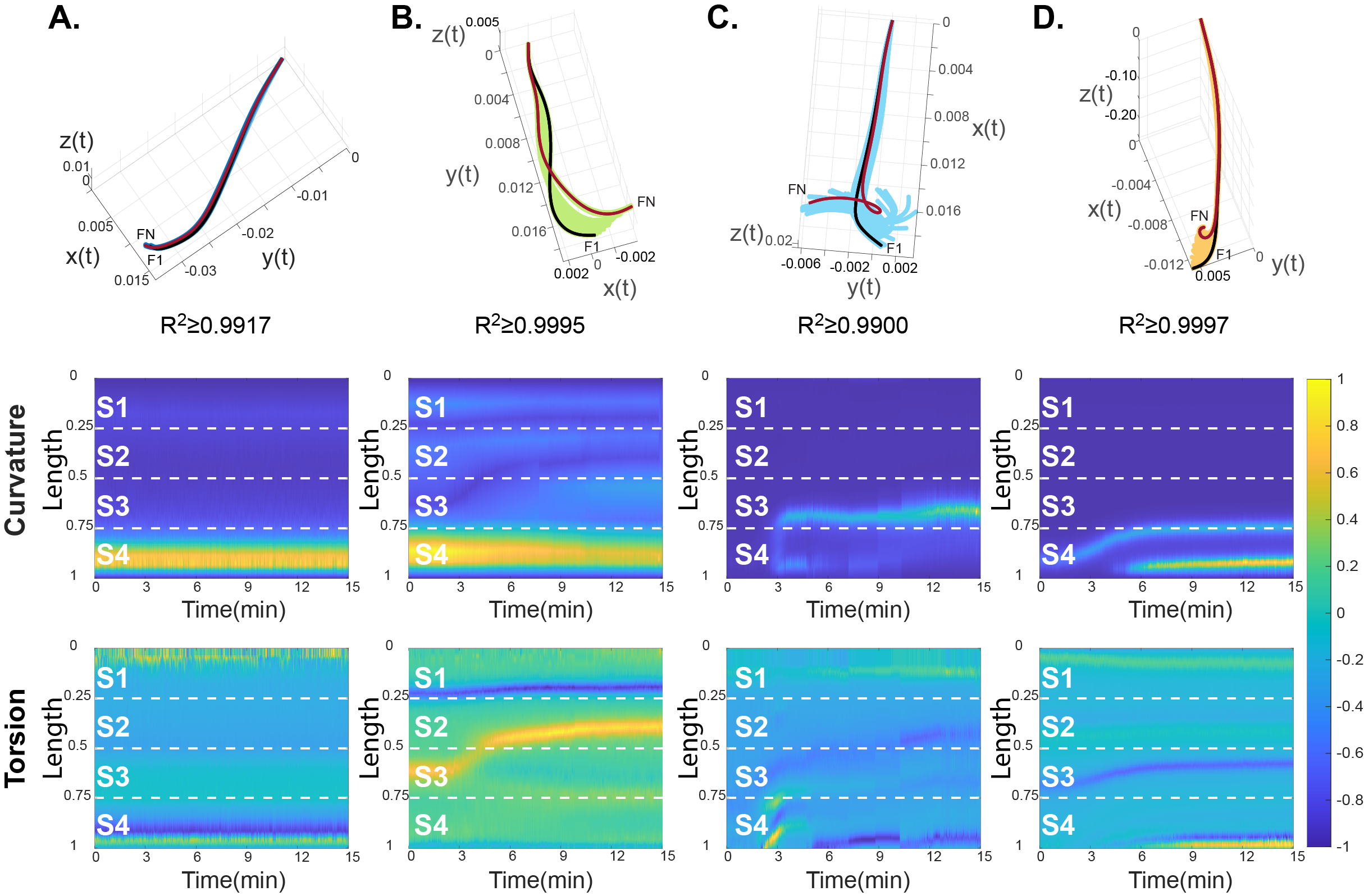}
    \caption{The comparison of four representative tendrils (shown by column) stimulated over the four different regions. The column A.B.C.D represents a tendril stimulated in region \textbf{S1}, \textbf{S2}, \textbf{S3}, \textbf{S4}, respectively. For each tendril, we show (top) the morphology evolution with the tendril in the first frame shown in black and the final frame shown in red; (middle) the normalized curvature (color bar) along the normalized tendril length; and (bottom) the normalized torsion (color bar) along the normalized tendril length.}  
    \label{fig:result_KT_Evolution}
\end{figure}

Figure \ref{fig:result_KT_Evolution} shows the evolution of four representative tendrils reconstructed with the proposed algorithm. In all cases, the tendrils exhibit morphological responses within 3 minutes of stimulus application. Except for the case $\textbf{S1}$ (Figure \ref{fig:result_KT_Evolution}A column), these responses produce significant deformations, as shown in the first row of Figure \ref{fig:result_KT_Evolution}.
For the case $\textbf{S2}$ (Figure \ref{fig:result_KT_Evolution}B column), there is a significant change in curvature within the stimulated area, accompanied by a large, dynamic alteration in torsion. Similar responses are seen for stimuli applied in the $\textbf{S3}$ and $\textbf{S4}$ regions. In particular, unlike previous results, localized stimulation not only affects the directly targeted area but also induces responses in adjacent regions of the tendril.

\subsection{Analysis of Tip Motion}
\begin{figure}
\centering
\includegraphics[width=\textwidth]{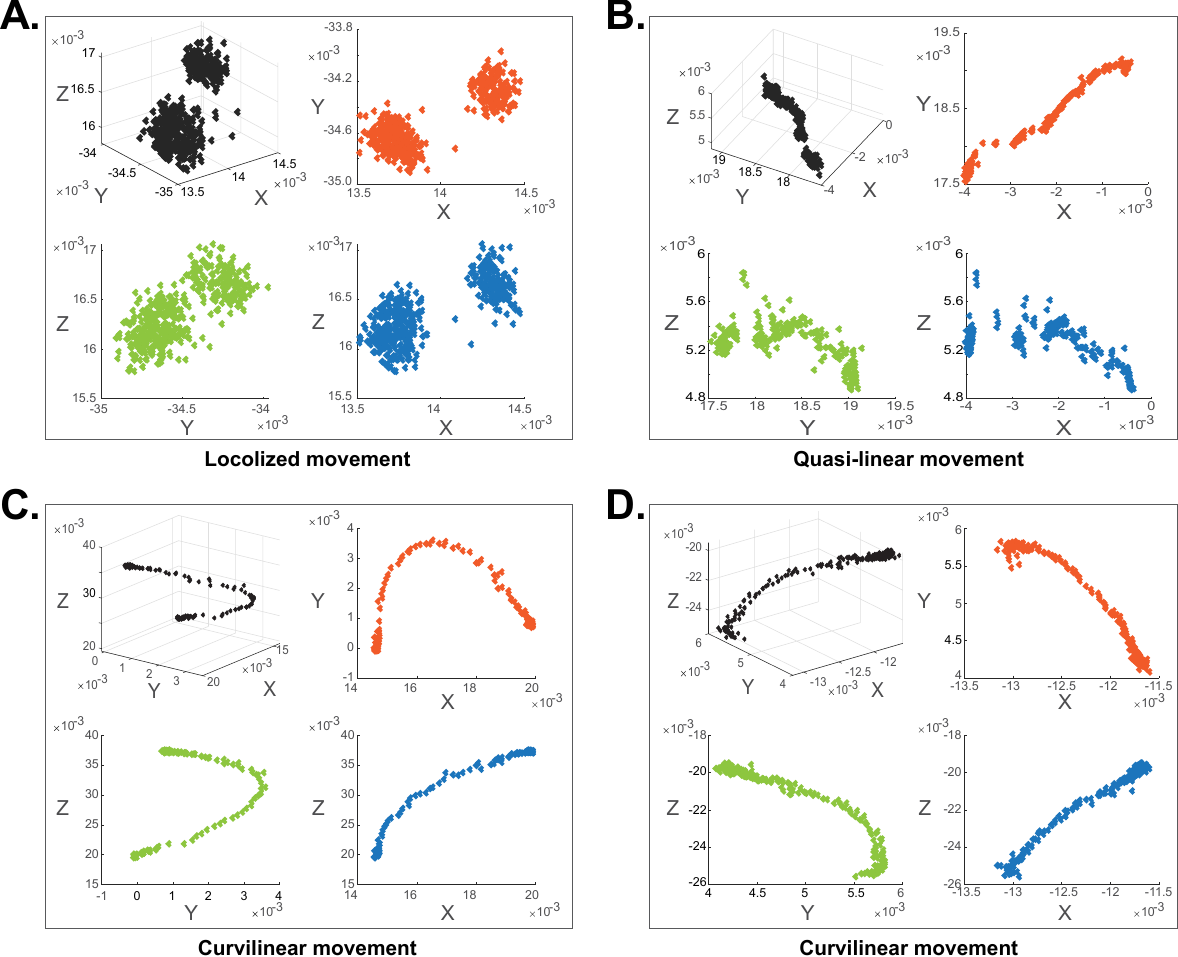}
\caption{3D tip movement of tendrils in black and the correspondent projections of trajectories to X-Y, Y-Z, X-Z planar, respectively colored in orange, green and blue.}
\label{fig:Tipmovement}
\end{figure}

Thanks to the proposed algorithm, we can better analyze the motion at the very apical extremity upon stimulation (Figure~\ref{fig:Tipmovement}). When the stimulus occurred in the \textbf{S1} region (Figure~\ref{fig:Tipmovement}A), the tip showed a stable movement around a central point, motivated by the stability of the shape that does not change significantly. The movement changes when the stimulus is applied to the \textbf{S2} region (Figure~\ref{fig:Tipmovement}B), showing a quasi-linear movement in space. For the last two cases (Figure~\ref{fig:Tipmovement}C and D), stimulus in regions \textbf{S3} and \textbf{S4}, the tip of the tendril moves following a curvilinear pattern spanning in large volume. The different tip movement patterns support our previous highlights: regions of the tendrils closer to the base result less active with respect to regions proximal to the tip. 

\section{Discussion}\label{section:Discussion}
In this work, we present a methodology to reconstruct and analyze plant tendril curling shape and movement based on stereo vision reconstruction and 3D piecewise clothoid curve fitting without need for marker-based features.
We used this approach to describe different 3D morphologies assumed by natural tendrils triggered by mechanical stimulation and analyze their shape evolution over time. This work provides a reliable (high accuracy, $R^2 \geq 0.99$) approach for reconstructing and modeling filamentous continuum structures. It demonstrates its applicability to the morphological description of tendrils of \textit{Passiflora cerulea} and contributed to analyzing the stimulus-response relationship in this model organ of climbing plants.
We found that tendrils respond to rub stimulation ($0.12~ N$ for $2 $~min) with a response time of about 3 $min$, and a final shape configuration stabilizing after $12 $~min on average. We could also identify regions with the highest responsiveness, which are located at the apical extremities of the organs.

From our results, we can infer information regarding the stiffness and sensing of these organs and draw specifications for the design of slender, filamentous, continuum, curling robots targeting anchoring or grasping through coiling deformations. For example, we can suggest a variable stiffness distribution of the structure going from stiffer to softer from base to tip and greater sensitivity to be displaced at the tip \cite{Weiqiang_2021}. 
In fact, we did not observe any meaningful response when stimulating the tendrils at the segment closer to the base. Also, the same segment was not showing a significant shape response when other segments were stimulated, while it was occurring for the other cases. These observations suggest that the basal part can have higher stiffness with respect to other locations, thus preventing any possible response. 
At the same time, low or no sensitivity can be inferred in the basal region since no response was activated in neighboring morphing regions. These features can enable more robust anchoring with the plant from the basal region while tightening to external support by morphing at the tip.

The proposed method is implemented based on post-stimulus reconstruction and a static geometrical representation of curling behavior, thus neglecting the time-dependent forces and dynamics during deformation. This limitation is particularly relevant for soft robotic applications requiring real-time actuation and force feedback. Future work should integrate our static geometric framework with dynamic continuum mechanics models, such as Cosserat rod theory \cite{Matteo_2023}, to capture time-varying deformations under external forces and internal actuation. Such integration would enable the reconstruction pipeline to serve as a component within closed-loop control architectures, where high-fidelity centerline representations provide observational constraints for physics-based simulations and inverse dynamics inform model-predictive control schemes. Extending the temporal resolution to capture transient dynamics would facilitate data-driven identification of material parameters and constitutive relationships, bridging the gap between kinematic shape description and mechanistic understanding of morphing behavior to design adaptive, tendril-inspired coiling robotic systems.

While demonstrated on \textit{Passiflora caerulea} tendrils, the proposed methodology applies to a broader class of filamentous continuum structures that exhibit comparable geometric characteristics. The principal requirements are: (1) high aspect ratios (length-to-diameter $>50:1$) where slender geometry justifies centerline-based representation, (2) ability to extract the skeleton of the slender structure and ensure the mapping of skeletal points between views for stereo correspondence, and (3) quasi-static or slowly evolving deformations amenable to discrete temporal sampling. These criteria are satisfied by diverse biological systems, including tendrils of other climbing species (e.g., \textit{Vitis vinifera}, \textit{Cucurbita}), plant roots, pollen tubes, fungal hyphae, and cellular protrusions, as well as engineered structures, such as soft robotic manipulators, surgical catheters, and flexible cables. The primary adaptations required would involve adjusting PWC penalty parameters to reflect different stiffness distributions and potentially incorporating domain-specific priors (e.g., gravitropic effects in roots, or pressure-driven deformation in catheters). Comprehensive validation across these diverse morphological systems remains an important direction for future investigation. In addition, future work could explore automation via Bayesian optimization or learning-based prediction of penalty values from geometric features to reduce computational overhead for real-time applications.

The observed increase in reconstruction and fitting errors over time (Tables 1–2) reflects error propagation inherent to our sequential processing pipeline. Since the first frame serves as the initialization reference, small uncertainties in feature localization propagate through subsequent frames as the tendril undergoes substantial morphological deformation. This temporal drift is compounded by increasing geometric complexity: as tendrils transition from nearly straight configurations to self-intersected curls, stereo correspondence becomes more challenging due to foreshortening effects and potential self-occlusions. Several mitigation strategies could reduce error accumulation, including periodic re-initialization at morphologically stable keyframes to prevent unbounded drift, applying bundle adjustment across sliding temporal windows to distribute errors more evenly, incorporating physics-based motion priors (e.g., Cosserat rod dynamics) to constrain biologically implausible deformations, and increasing temporal sampling frequency during rapid transitions to improve feature tracking reliability, which also remain for future work.

\bmhead{Acknowledgments}
We acknowledge the support and funding from the European Union’s Horizon 2020 Research and Innovation Program under Grant Agreement No.824074 (GrowBot).

\bmhead{Data Availability}
The anonymized datasets and source code scripts supporting this study are publicly available under an open-source license at GitHub (\url{https://github.com/jiefan0414/IJCV25-3d_Tendril-like_shape_reconstruction.git}). Additional datasets collected and analyzed during the current study are available from the corresponding author upon reasonable request.

\input{main.bbl}

\end{document}

%% file: main.bbl

%% file: main.bbl
\begin{thebibliography}{34}
\ifx \bisbn   \undefined \def \bisbn  #1{ISBN #1}\fi
\ifx \binits  \undefined \def \binits#1{#1}\fi
\ifx \bauthor  \undefined \def \bauthor#1{#1}\fi
\ifx \batitle  \undefined \def \batitle#1{#1}\fi
\ifx \bjtitle  \undefined \def \bjtitle#1{#1}\fi
\ifx \bvolume  \undefined \def \bvolume#1{\textbf{#1}}\fi
\ifx \byear  \undefined \def \byear#1{#1}\fi
\ifx \bissue  \undefined \def \bissue#1{#1}\fi
\ifx \bfpage  \undefined \def \bfpage#1{#1}\fi
\ifx \blpage  \undefined \def \blpage #1{#1}\fi
\ifx \burl  \undefined \def \burl#1{\textsf{#1}}\fi
\ifx \doiurl  \undefined \def \doiurl#1{\url{https://doi.org/#1}}\fi
\ifx \betal  \undefined \def \betal{\textit{et al.}}\fi
\ifx \binstitute  \undefined \def \binstitute#1{#1}\fi
\ifx \binstitutionaled  \undefined \def \binstitutionaled#1{#1}\fi
\ifx \bctitle  \undefined \def \bctitle#1{#1}\fi
\ifx \beditor  \undefined \def \beditor#1{#1}\fi
\ifx \bpublisher  \undefined \def \bpublisher#1{#1}\fi
\ifx \bbtitle  \undefined \def \bbtitle#1{#1}\fi
\ifx \bedition  \undefined \def \bedition#1{#1}\fi
\ifx \bseriesno  \undefined \def \bseriesno#1{#1}\fi
\ifx \blocation  \undefined \def \blocation#1{#1}\fi
\ifx \bsertitle  \undefined \def \bsertitle#1{#1}\fi
\ifx \bsnm \undefined \def \bsnm#1{#1}\fi
\ifx \bsuffix \undefined \def \bsuffix#1{#1}\fi
\ifx \bparticle \undefined \def \bparticle#1{#1}\fi
\ifx \barticle \undefined \def \barticle#1{#1}\fi
\bibcommenthead
\ifx \bconfdate \undefined \def \bconfdate #1{#1}\fi
\ifx \botherref \undefined \def \botherref #1{#1}\fi
\ifx \url \undefined \def \url#1{\textsf{#1}}\fi
\ifx \bchapter \undefined \def \bchapter#1{#1}\fi
\ifx \bbook \undefined \def \bbook#1{#1}\fi
\ifx \bcomment \undefined \def \bcomment#1{#1}\fi
\ifx \oauthor \undefined \def \oauthor#1{#1}\fi
\ifx \citeauthoryear \undefined \def \citeauthoryear#1{#1}\fi
\ifx \endbibitem  \undefined \def \endbibitem {}\fi
\ifx \bconflocation  \undefined \def \bconflocation#1{#1}\fi
\ifx \arxivurl  \undefined \def \arxivurl#1{\textsf{#1}}\fi
\csname PreBibitemsHook\endcsname

\bibitem[\protect\citeauthoryear{Laschi et~al.}{2016}]{laschi_2016soft}
\begin{barticle}
\bauthor{\bsnm{Laschi}, \binits{C.}},
\bauthor{\bsnm{Mazzolai}, \binits{B.}},
\bauthor{\bsnm{Cianchetti}, \binits{M.}}:
\batitle{Soft robotics: Technologies and systems pushing the boundaries of robot abilities}.
\bjtitle{Science Robotics}
\bvolume{1}(\bissue{1}),
\bfpage{3690}
(\byear{2016})
\end{barticle}
\endbibitem

\bibitem[\protect\citeauthoryear{Laschi et~al.}{2009}]{Laschi_2009}
\begin{botherref}
\oauthor{\bsnm{Laschi}, \binits{C.}},
\oauthor{\bsnm{Mazzolai}, \binits{B.}},
\oauthor{\bsnm{Mattoli}, \binits{V.}},
\oauthor{\bsnm{Cianchetti}, \binits{M.}},
\oauthor{\bsnm{Dario}, \binits{P.}}:
Design of a biomimetic robotic octopus arm.
Bioinspiration {\&} Biomimetics
(2009)
\end{botherref}
\endbibitem

\bibitem[\protect\citeauthoryear{Mazzolai et~al.}{2019}]{mazzolai2019octopus}
\begin{barticle}
\bauthor{\bsnm{Mazzolai}, \binits{B.}},
\bauthor{\bsnm{Mondini}, \binits{A.}},
\bauthor{\bsnm{Tramacere}, \binits{F.}},
\bauthor{\bsnm{Riccomi}, \binits{G.}},
\bauthor{\bsnm{Sadeghi}, \binits{A.}},
\bauthor{\bsnm{Giordano}, \binits{G.}},
\bauthor{\bsnm{Del~Dottore}, \binits{E.}},
\bauthor{\bsnm{Scaccia}, \binits{M.}},
\bauthor{\bsnm{Zampato}, \binits{M.}},
\bauthor{\bsnm{Carminati}, \binits{S.}}:
\batitle{Octopus-inspired soft arm with suction cups for enhanced grasping tasks in confined environments}.
\bjtitle{Advanced Intelligent Systems}
\bvolume{1}(\bissue{6}),
\bfpage{1900041}
(\byear{2019})
\end{barticle}
\endbibitem

\bibitem[\protect\citeauthoryear{Kim et~al.}{2013}]{kim_2013soft}
\begin{barticle}
\bauthor{\bsnm{Kim}, \binits{S.}},
\bauthor{\bsnm{Laschi}, \binits{C.}},
\bauthor{\bsnm{Trimmer}, \binits{B.}}:
\batitle{Soft robotics: a bioinspired evolution in robotics}.
\bjtitle{Trends in biotechnology}
\bvolume{31}(\bissue{5}),
\bfpage{287}--\blpage{294}
(\byear{2013})
\end{barticle}
\endbibitem

\bibitem[\protect\citeauthoryear{Isnard and Silk}{2009}]{Isnard_2009}
\begin{barticle}
\bauthor{\bsnm{Isnard}, \binits{S.}},
\bauthor{\bsnm{Silk}, \binits{W.}}:
\batitle{Moving with climbing plants from charles darwin's time into the 21st century}.
\bjtitle{American journal of botany}
\bvolume{96},
\bfpage{1205}--\blpage{21}
(\byear{2009})
\end{barticle}
\endbibitem

\bibitem[\protect\citeauthoryear{Burris et~al.}{2017}]{Burris_2017}
\begin{botherref}
\oauthor{\bsnm{Burris}, \binits{J.}},
\oauthor{\bsnm{Lenaghan}, \binits{S.}},
\oauthor{\bsnm{Stewart}, \binits{C.}}:
Climbing plants: attachment adaptations and bioinspired innovations.
Plant Cell Reports
(2017)
\end{botherref}
\endbibitem

\bibitem[\protect\citeauthoryear{Chehab et~al.}{2008}]{Chehab_2008}
\begin{barticle}
\bauthor{\bsnm{Chehab}, \binits{E.W.}},
\bauthor{\bsnm{Eich}, \binits{E.}},
\bauthor{\bsnm{Braam}, \binits{J.}}:
\batitle{{Thigmomorphogenesis: a complex plant response to mechano-stimulation}}.
\bjtitle{Journal of Experimental Botany}
\bvolume{60}(\bissue{1}),
\bfpage{43}--\blpage{56}
(\byear{2008})
\end{barticle}
\endbibitem

\bibitem[\protect\citeauthoryear{Jaffe and Galston}{1968}]{Jaffe_1968}
\begin{barticle}
\bauthor{\bsnm{Jaffe}, \binits{M.J.}},
\bauthor{\bsnm{Galston}, \binits{A.W.}}:
\batitle{The physiology of tendrils}.
\bjtitle{Annual Review of Plant Physiology}
\bvolume{19}(\bissue{1}),
\bfpage{417}--\blpage{434}
(\byear{1968})
\end{barticle}
\endbibitem

\bibitem[\protect\citeauthoryear{Sousa-Baena et~al.}{2018}]{Baena_2018_2}
\begin{botherref}
\oauthor{\bsnm{Sousa-Baena}, \binits{M.S.}},
\oauthor{\bsnm{Lohmann}, \binits{L.G.}},
\oauthor{\bsnm{Hernandes-Lopes}, \binits{J.}},
\oauthor{\bsnm{Sinha}, \binits{N.R.}}:
The molecular control of tendril development in angiosperms.
New Phytologist
\textbf{218}(3)
(2018)
\end{botherref}
\endbibitem

\bibitem[\protect\citeauthoryear{Vidoni et~al.}{2015}]{Vidoni_2015}
\begin{botherref}
\oauthor{\bsnm{Vidoni}, \binits{R.}},
\oauthor{\bsnm{Mimmo}, \binits{T.}},
\oauthor{\bsnm{Pandolfi}, \binits{C.}}:
Tendril-based climbing plants to model, simulate and create bio-inspired robotic systems.
Journal of Bionic Engineering
\textbf{12}
(2015)
\end{botherref}
\endbibitem

\bibitem[\protect\citeauthoryear{Keklikoglou et~al.}{2021}]{Keklikoglou_2021}
\begin{botherref}
\oauthor{\bsnm{Keklikoglou}, \binits{K.}},
\oauthor{\bsnm{Arvanitidis}, \binits{C.}},
\oauthor{\bsnm{Chatzigeorgiou}, \binits{G.}},
\oauthor{\bsnm{Chatzinikolaou}, \binits{E.}},
\oauthor{\bsnm{Karagiannidis}, \binits{E.}},
\oauthor{\bsnm{Koletsa}, \binits{T.}},
\oauthor{\bsnm{Magoulas}, \binits{A.}},
\oauthor{\bsnm{Makris}, \binits{K.}},
\oauthor{\bsnm{Mavrothalassitis}, \binits{G.}},
\oauthor{\bsnm{Papanagnou}, \binits{E.-D.}},
\oauthor{\bsnm{Papazoglou}, \binits{A.S.}},
\oauthor{\bsnm{Pavloudi}, \binits{C.}},
\oauthor{\bsnm{Trougakos}, \binits{I.P.}},
\oauthor{\bsnm{Vasileiadou}, \binits{K.}},
\oauthor{\bsnm{Vogiatzi}, \binits{A.}}:
Micro-ct for biological and biomedical studies: A comparison of imaging techniques.
Journal of Imaging
\textbf{7}(9)
(2021)
\end{botherref}
\endbibitem

\bibitem[\protect\citeauthoryear{Giordano et~al.}{2021}]{giordano2021perspective}
\begin{barticle}
\bauthor{\bsnm{Giordano}, \binits{G.}},
\bauthor{\bsnm{Carlotti}, \binits{M.}},
\bauthor{\bsnm{Mazzolai}, \binits{B.}}:
\batitle{A perspective on cephalopods mimicry and bioinspired technologies toward proprioceptive autonomous soft robots}.
\bjtitle{Advanced Materials Technologies}
\bvolume{6}(\bissue{12}),
\bfpage{2100437}
(\byear{2021})
\end{barticle}
\endbibitem

\bibitem[\protect\citeauthoryear{Moeslund et~al.}{2006}]{Moeslund_2006}
\begin{barticle}
\bauthor{\bsnm{Moeslund}, \binits{T.B.}},
\bauthor{\bsnm{Hilton}, \binits{A.}},
\bauthor{\bsnm{Krüger}, \binits{V.}}:
\batitle{A survey of advances in vision-based human motion capture and analysis}.
\bjtitle{Computer Vision and Image Understanding}
\bvolume{104}(\bissue{2}),
\bfpage{90}--\blpage{126}
(\byear{2006})
\doiurl{10.1016/j.cviu.2006.08.002} .
\bcomment{Special Issue on Modeling People: Vision-based understanding of a person’s shape, appearance, movement and behaviour}
\end{barticle}
\endbibitem

\bibitem[\protect\citeauthoryear{Hartley and Zisserman}{2004}]{Hartley_2004}
\begin{bbook}
\bauthor{\bsnm{Hartley}, \binits{R.}},
\bauthor{\bsnm{Zisserman}, \binits{A.}}:
\bbtitle{Multiple View Geometry in Computer Vision},
\bedition{2}nd edn.
\bpublisher{Cambridge University Press},
\blocation{Cambridge}
(\byear{2004})
\end{bbook}
\endbibitem

\bibitem[\protect\citeauthoryear{Mildenhall et~al.}{2021}]{Mildenhall_2021_NeRF}
\begin{barticle}
\bauthor{\bsnm{Mildenhall}, \binits{B.}},
\bauthor{\bsnm{Srinivasan}, \binits{P.P.}},
\bauthor{\bsnm{Tancik}, \binits{M.}},
\bauthor{\bsnm{Barron}, \binits{J.T.}},
\bauthor{\bsnm{Ramamoorthi}, \binits{R.}},
\bauthor{\bsnm{Ng}, \binits{R.}}:
\batitle{Nerf: representing scenes as neural radiance fields for view synthesis}.
\bjtitle{Commun. ACM}
\bvolume{65}(\bissue{1}),
\bfpage{99}--\blpage{106}
(\byear{2021})
\doiurl{10.1145/3503250}
\end{barticle}
\endbibitem

\bibitem[\protect\citeauthoryear{Kerbl et~al.}{2023}]{Kerbl_2023_gaussians}
\begin{botherref}
\oauthor{\bsnm{Kerbl}, \binits{B.}},
\oauthor{\bsnm{Kopanas}, \binits{G.}},
\oauthor{\bsnm{Leimk{\"u}hler}, \binits{T.}},
\oauthor{\bsnm{Drettakis}, \binits{G.}}:
3d gaussian splatting for real-time radiance field rendering.
ACM Transactions on Graphics
\textbf{42}(4)
(2023)
\end{botherref}
\endbibitem

\bibitem[\protect\citeauthoryear{Wolf et~al.}{2024}]{wolf_2024_gsmesh_eccv}
\begin{bchapter}
\bauthor{\bsnm{Wolf}, \binits{Y.}},
\bauthor{\bsnm{Bracha}, \binits{A.}},
\bauthor{\bsnm{Kimmel}, \binits{R.}}:
\bctitle{{GS}2{M}esh: Surface reconstruction from {G}aussian splatting via novel stereo views}.
In: \bbtitle{European Conference on Computer Vision (ECCV)}
(\byear{2024})
\end{bchapter}
\endbibitem

\bibitem[\protect\citeauthoryear{Wang et~al.}{2021}]{wang2021neus}
\begin{botherref}
\oauthor{\bsnm{Wang}, \binits{P.}},
\oauthor{\bsnm{Liu}, \binits{L.}},
\oauthor{\bsnm{Liu}, \binits{Y.}},
\oauthor{\bsnm{Theobalt}, \binits{C.}},
\oauthor{\bsnm{Komura}, \binits{T.}},
\oauthor{\bsnm{Wang}, \binits{W.}}:
Neus: Learning neural implicit surfaces by volume rendering for multi-view reconstruction.
NeurIPS
(2021)
\end{botherref}
\endbibitem

\bibitem[\protect\citeauthoryear{Yu et~al.}{2022}]{Yu2022MonoSDF}
\begin{botherref}
\oauthor{\bsnm{Yu}, \binits{Z.}},
\oauthor{\bsnm{Peng}, \binits{S.}},
\oauthor{\bsnm{Niemeyer}, \binits{M.}},
\oauthor{\bsnm{Sattler}, \binits{T.}},
\oauthor{\bsnm{Geiger}, \binits{A.}}:
Monosdf: Exploring monocular geometric cues for neural implicit surface reconstruction.
Advances in Neural Information Processing Systems (NeurIPS)
(2022)
\end{botherref}
\endbibitem

\bibitem[\protect\citeauthoryear{Yang et~al.}{2025}]{yang2025sgcr}
\begin{bchapter}
\bauthor{\bsnm{Yang}, \binits{X.}},
\bauthor{\bsnm{Ji}, \binits{D.}},
\bauthor{\bsnm{Li}, \binits{Y.}},
\bauthor{\bsnm{Guo}, \binits{J.}},
\bauthor{\bsnm{Guo}, \binits{Y.}},
\bauthor{\bsnm{Xie}, \binits{J.}}:
\bctitle{Sgcr: Spherical gaussians for efficient 3d curve reconstruction}.
In: \bbtitle{Proceedings of the IEEE/CVF Conference on Computer Vision and Pattern Recognition (CVPR)}
(\byear{2025})
\end{bchapter}
\endbibitem

\bibitem[\protect\citeauthoryear{Robert J.~Webster and Jones}{2010}]{Webster_2010}
\begin{barticle}
\bauthor{\bsnm{Robert J.~Webster}, \binits{I.}},
\bauthor{\bsnm{Jones}, \binits{B.A.}}:
\batitle{Design and kinematic modeling of constant curvature continuum robots: A review}.
\bjtitle{The International Journal of Robotics Research}
\bvolume{29}(\bissue{13}),
\bfpage{1661}--\blpage{1683}
(\byear{2010})
\end{barticle}
\endbibitem

\bibitem[\protect\citeauthoryear{Trivedi et~al.}{2008}]{Trivedi_2008}
\begin{barticle}
\bauthor{\bsnm{Trivedi}, \binits{D.}},
\bauthor{\bsnm{Rahn}, \binits{C.D.}},
\bauthor{\bsnm{Kier}, \binits{W.M.}},
\bauthor{\bsnm{Walker}, \binits{I.D.}}:
\batitle{Soft robotics: Biological inspiration, state of the art, and future research}.
\bjtitle{Applied Bionics and Biomechanics}
\bvolume{5}(\bissue{3}),
\bfpage{520417}
(\byear{2008})
\doiurl{10.1080/11762320802557865}
{\href{https://arxiv.org/abs/https://onlinelibrary.wiley.com/doi/pdf/10.1080/11762320802557865}{{https://onlinelibrary.wiley.com/doi/pdf/10.1080/11762320802557865}}}
\end{barticle}
\endbibitem

\bibitem[\protect\citeauthoryear{Fan et~al.}{2020a}]{Fan_2020_1}
\begin{bchapter}
\bauthor{\bsnm{Fan}, \binits{J.}},
\bauthor{\bsnm{Del~Dottore}, \binits{E.}},
\bauthor{\bsnm{Visentin}, \binits{F.}},
\bauthor{\bsnm{Mazzolai}, \binits{B.}}:
\bctitle{Image-based approach to reconstruct curling in continuum structures}.
In: \bbtitle{2020 3rd IEEE International Conference on Soft Robotics (RoboSoft)},
pp. \bfpage{544}--\blpage{549}
(\byear{2020})
\end{bchapter}
\endbibitem

\bibitem[\protect\citeauthoryear{Fan et~al.}{2020b}]{Fan_2020_2}
\begin{bchapter}
\bauthor{\bsnm{Fan}, \binits{J.}},
\bauthor{\bsnm{Visentin}, \binits{F.}},
\bauthor{\bsnm{Del~Dottore}, \binits{E.}},
\bauthor{\bsnm{Mazzolai}, \binits{B.}}:
\bctitle{An image-based method for the morphological analysis of tendrils with 2d piece-wise clothoid approximation model}.
In: \bbtitle{Biomimetic and Biohybrid Systems},
pp. \bfpage{80}--\blpage{91}.
\bpublisher{Springer},
\blocation{Cham}
(\byear{2020})
\end{bchapter}
\endbibitem

\bibitem[\protect\citeauthoryear{Jaffe}{1970}]{Jaffe_1970}
\begin{barticle}
\bauthor{\bsnm{Jaffe}, \binits{M.J.}}:
\batitle{Physiological studies on pea tendrils: Vi. the characteristics of sensory perception and transduction}.
\bjtitle{Plant Physiology}
\bvolume{45}(\bissue{6}),
\bfpage{756}--\blpage{760}
(\byear{1970})
\end{barticle}
\endbibitem

\bibitem[\protect\citeauthoryear{Jaffe and Galston}{1966}]{Jaffe_1966}
\begin{barticle}
\bauthor{\bsnm{Jaffe}, \binits{M.J.}},
\bauthor{\bsnm{Galston}, \binits{A.W.}}:
\batitle{{Physiological Studies on Pea Tendrils. I. Growth and Coiling Following Mechanical Stimulation}}.
\bjtitle{Plant Physiology}
\bvolume{41}(\bissue{6}),
\bfpage{1014}--\blpage{1025}
(\byear{1966})
\end{barticle}
\endbibitem

\bibitem[\protect\citeauthoryear{Kovesi}{}]{KovesiMATLABCode}
\begin{botherref}
\oauthor{\bsnm{Kovesi}, \binits{P.D.}}:
{MATLAB} and {Octave} Functions for Computer Vision and Image Processing.
Available from: $<$https://www.peterkovesi.com/matlabfns/$>$
\end{botherref}
\endbibitem

\bibitem[\protect\citeauthoryear{Thomas and Heikki}{1994}]{Eiter_1994}
\begin{bchapter}
\bauthor{\bsnm{Thomas}, \binits{E.}},
\bauthor{\bsnm{Heikki}, \binits{M.}}:
\bctitle{Computing discrete fréchet distance.}
(\byear{1994})
\end{bchapter}
\endbibitem

\bibitem[\protect\citeauthoryear{Harary and Tal}{2010}]{Harary_2010}
\begin{bchapter}
\bauthor{\bsnm{Harary}, \binits{G.}},
\bauthor{\bsnm{Tal}, \binits{A.}}:
\bctitle{3d euler spirals for 3d curve completion}.
In: \bbtitle{Proceedings of the Twenty-Sixth Annual Symposium on Computational Geometry}.
\bsertitle{SoCG '10},
pp. \bfpage{393}--\blpage{402}.
\bpublisher{Association for Computing Machinery},
\blocation{New York, NY, USA}
(\byear{2010})
\end{bchapter}
\endbibitem

\bibitem[\protect\citeauthoryear{Hu et~al.}{2011}]{Shuangwei_2011}
\begin{barticle}
\bauthor{\bsnm{Hu}, \binits{S.}},
\bauthor{\bsnm{Lundgren}, \binits{M.}},
\bauthor{\bsnm{Niemi}, \binits{A.J.}}:
\batitle{Discrete frenet frame, inflection point solitons, and curve visualization with applications to folded proteins}.
\bjtitle{Phys. Rev. E}
\bvolume{83},
\bfpage{061908}
(\byear{2011})
\end{barticle}
\endbibitem

\bibitem[\protect\citeauthoryear{Lim and Han}{2017}]{Lim_2017}
\begin{botherref}
\oauthor{\bsnm{Lim}, \binits{S.}},
\oauthor{\bsnm{Han}, \binits{S.}}:
Helical extension method for solving the natural equation of a space curve
\textbf{5}(3),
035002
(2017)
\end{botherref}
\endbibitem

\bibitem[\protect\citeauthoryear{Bouaziz et~al.}{2016}]{Bouaziz_2016}
\begin{bchapter}
\bauthor{\bsnm{Bouaziz}, \binits{S.}},
\bauthor{\bsnm{Tagliasacchi}, \binits{A.}},
\bauthor{\bsnm{Li}, \binits{H.}},
\bauthor{\bsnm{Pauly}, \binits{M.}}:
\bctitle{Modern techniques and applications for real-time non-rigid registration}.
In: \bbtitle{SIGGRAPH ASIA 2016 Courses}.
\bsertitle{SA '16}.
\bpublisher{Association for Computing Machinery},
\blocation{New York, NY, USA}
(\byear{2016})
\end{bchapter}
\endbibitem

\bibitem[\protect\citeauthoryear{Dou et~al.}{2021}]{Weiqiang_2021}
\begin{barticle}
\bauthor{\bsnm{Dou}, \binits{W.}},
\bauthor{\bsnm{Zhong}, \binits{G.}},
\bauthor{\bsnm{Cao}, \binits{J.}},
\bauthor{\bsnm{Shi}, \binits{Z.}},
\bauthor{\bsnm{Peng}, \binits{B.}},
\bauthor{\bsnm{Jiang}, \binits{L.}}:
\batitle{Soft robotic manipulators: Designs, actuation, stiffness tuning, and sensing}.
\bjtitle{Advanced Materials Technologies}
\bvolume{6}(\bissue{9}),
\bfpage{2100018}
(\byear{2021})
\doiurl{10.1002/admt.202100018}
{\href{https://arxiv.org/abs/https://advanced.onlinelibrary.wiley.com/doi/pdf/10.1002/admt.202100018}{{https://advanced.onlinelibrary.wiley.com/doi/pdf/10.1002/admt.202100018}}}
\end{barticle}
\endbibitem

\bibitem[\protect\citeauthoryear{Russo et~al.}{2023}]{Matteo_2023}
\begin{barticle}
\bauthor{\bsnm{Russo}, \binits{M.}},
\bauthor{\bsnm{Sadati}, \binits{S.M.H.}},
\bauthor{\bsnm{Dong}, \binits{X.}},
\bauthor{\bsnm{Mohammad}, \binits{A.}},
\bauthor{\bsnm{Walker}, \binits{I.D.}},
\bauthor{\bsnm{Bergeles}, \binits{C.}},
\bauthor{\bsnm{Xu}, \binits{K.}},
\bauthor{\bsnm{Axinte}, \binits{D.A.}}:
\batitle{Continuum robots: An overview}.
\bjtitle{Advanced Intelligent Systems}
\bvolume{5}(\bissue{5}),
\bfpage{2200367}
(\byear{2023})
\doiurl{10.1002/aisy.202200367}
{\href{https://arxiv.org/abs/https://advanced.onlinelibrary.wiley.com/doi/pdf/10.1002/aisy.202200367}{{https://advanced.onlinelibrary.wiley.com/doi/pdf/10.1002/aisy.202200367}}}
\end{barticle}
\endbibitem

\end{thebibliography}
